\documentclass[journal]{IEEEtran}
\makeatletter
\def\endthebibliography{%
  \def\@noitemerr{\@latex@warning{Empty `thebibliography' environment}}%
  \endlist
}
\makeatother

\ifCLASSINFOpdf
\else
\fi
\usepackage{cite}
\usepackage{amsmath,amssymb,amsfonts}
\usepackage{algorithmic}
\usepackage{graphicx}
\usepackage{textcomp}
\def\BibTeX{{\rm B\kern-.05em{\sc i\kern-.025em b}\kern-.08em
    T\kern-.1667em\lower.7ex\hbox{E}\kern-.125emX}}
\usepackage{amsmath}
\usepackage[english]{babel} % English language/hyphenation
\usepackage{amsmath,amsfonts,amsthm,bm} % Math packages
\makeatletter
\def\dual#1{\expandafter\dual@aux#1\@nil}
\def\dual@aux#1/#2\@nil{\begin{tabular}{@{}c@{}}#1\\#2\end{tabular}}
\makeatother
\usepackage{amsmath}
\usepackage{mathtools}
\usepackage{multirow}
\usepackage{footnote}
\makesavenoteenv{tabular}
\makesavenoteenv{table}
\usepackage{mathtools}
\usepackage{caption}
\usepackage{cite}
\usepackage{textcase}
\usepackage[tablename=TABLE]{caption}
\usepackage{lipsum}% http://ctan.org/pkg/lipsum

\usepackage{array}
\newcolumntype{P}[1]{>{\centering\arraybackslash}p{#1}}

\usepackage{color}
\usepackage[table, dvipsnames]{xcolor}

\usepackage[font=scriptsize]{caption}

\usepackage{url}
\usepackage[belowskip=-15pt,aboveskip=0pt]{caption}
\usepackage[skip=8pt]{caption} % example skip set to 2pt
\usepackage{epstopdf} 
\usepackage{caption}
\usepackage{cite}
\usepackage{textcase}
\usepackage[tablename=TABLE]{caption}
\usepackage[font=scriptsize]{caption}
\usepackage{mathtools}
\usepackage{bm}
\usepackage{amsmath}
\newcommand{\vect}[1]{\mathbf{#1}} % or \bm{#1}
\DeclarePairedDelimiter\abs{\lvert}{\rvert}%
\usepackage{amssymb}
\usepackage{amsfonts}
\usepackage{amsmath}
\usepackage{amsfonts}
\usepackage{amssymb}
\usepackage{graphicx,subcaption}
\usepackage{subcaption}
\usepackage{titlesec}
\titlespacing*{\section}{0pc}{0.5pc}{0.5pc}
\titlespacing*{\subsection} {0pc}{0.3pc}{0.3pc}

\IEEEaftertitletext{\vspace{-2.5\baselineskip}}
\usepackage{mathtools, nccmath}

\usepackage{enumitem}
\setlist{nolistsep,topsep=1pt}

\begin{document}
%\raggedbottom
\title{Multimodal Deep Learning Framework for Image Popularity Prediction on Social Media
}
%Wen-Huang~Cheng,~\IEEEmembership{Fellow,~OSA,}

\author{Fatma~S.~Abousaleh,
        Wen-Huang~Cheng,~Neng-Hao~Yu,~and Yu~Tsao,~\IEEEmembership{Senior Member,~IEEE,}
%        , Hong-Yuan Mark Liao,~\IEEEmembership{Fellow,~IEEE}% <-this % stops a space

\thanks{(Corresponding author:Yu Tsao)

Fatma S. Abousaleh is with Social Networks and Human-Centered Computing Program, Taiwan International Graduate Program, Institute of Information Science and Research Center for Information Technology Innovation,  Academia Sinica, Taipei 11529, Taiwan, and also with Department of Computer Science, National Chengchi University, Taipei 11605, Taiwan. (e-mail: fatma@iis.sinica.edu.tw).}% <-this % stops a space

\thanks{Wen-Huang Cheng is with the Institute of Electronics, Department of Electronics Engineering, National Chiao Tung University, Hsinchu 30010, Taiwan. (e-mail: whcheng@nctu.edu.tw).  }% <-this % stops a space
\thanks{Neng-Hao Yu is with the College of Design, Department of Design, National Taiwan University of Science and Technology, Taipei 10607, Taiwan. (e-mail: jonesyu@mail.ntust.edu.tw).  }% <-this % stops a space
\thanks{Yu Tsao is with the Research Center for Information Technology Innovation, Academia Sinica, Taipei 11529, Taiwan. (e-mail: yu.tsao@citi.sinica.edu.tw).  }% <-this % stops a space
%\thanks{Hong-Yuan Mark Liao is with the Institute of Information Science, Academia Sinica, Taipei 11529, Taiwan. (e-mail: liao@iis.sinica.edu.tw).  }% <-this % stops a space

 }

\maketitle
%\vspace*{-1.5cm}

\begin{abstract}
Billions of photos are uploaded to the web daily through various types of social networks. Some of these images receive millions of views and become popular, whereas others remain completely unnoticed. This raises the problem of predicting image popularity on social media. The popularity of an image can be affected by several factors, such as visual content, aesthetic quality, user, post metadata, and time. Thus, considering all these factors is essential for accurately predicting image popularity. In addition, the efficiency of the predictive model also plays a crucial role. In this study, motivated by multimodal learning, which uses information from various modalities, and the current success of convolutional neural networks (CNNs) in various fields, we propose a deep learning model, called visual-social convolutional neural network (VSCNN), which predicts the popularity of a posted image by incorporating various types of visual and social features into a unified network model. VSCNN first learns to extract high-level representations from the input visual and social features by utilizing two individual CNNs. The outputs of these two networks are then fused into a joint network to estimate the popularity score in the output layer. We assess the performance of the proposed method by conducting extensive experiments on a dataset of approximately 432K images posted on Flickr. The simulation results demonstrate that the proposed VSCNN model significantly outperforms state-of-the-art models, with a relative improvement of greater than 2.33\%, 7.59\%, and 14.16\% in terms of Spearman's Rho, mean absolute error, and mean squared error, respectively.

%comprehensive,large-scale

%The simulation results demonstrate that the proposed VSCNN model significantly outperforms six state-of-the-art baseline models and successfully achieve $0.9014$, $0.73$, and $0.97$ in terms of Spearman's rank correlation coefficient (Spearman's Rho), Mean Absolute Error (MAE), and Mean Squared Error (MSE), respectively. 

%The simulation results demonstrate that the proposed VSCNN model significantly outperforms six state-of-the-art baseline models, with a relative performance improvement of 2.33\% (in SCNN) in terms of Spearman's rank correlation coefficient (Spearman's Rho), and with decreases of 7.59\%  and 14.16\% in terms of Mean Absolute Error (MAE), and Mean Squared Error (MSE), respectively.

%The simulation results demonstrate that the proposed VSCNN model significantly outperforms six state-of-the-art baseline models, with a relative performance improvement of more than 2.33\%, 7.59\%, and 14.16\% in terms of Spearman's rank correlation coefficient, mean absolute error, and mean squared error, respectively. 
\end{abstract}


% \fi
\IEEEpeerreviewmaketitle

\section{Introduction}
\IEEEPARstart{S}{ocial} media websites (e.g., Flickr, Twitter, and Facebook) allow users to create and share content (e.g., by liking, commenting, or viewing). Consequently, social media platforms have become an inseparable part of our daily lives, with significant social content  generated on these platforms. The explosive growth of social media content (i.e., texts, images, audios, and videos) and the interactive behavior between web users result in that only a small portion of online social content attracts significant attention and becomes popular, whereas its vast majority either receives little attention or is entirely overlooked. Therefore, extensive efforts have been expended in the past few years to predict social media content popularity, understand its variation, and evaluate its growth \cite{ Ref1,niu2012predicting, Ref6, Ref7, Ref8, Ref9, Ref13}. This popularity reflects user interests and provides opportunities to understand user interaction with online content, as well as information diffusion through social media websites. Hence, an accurate popularity prediction of online content may improve user experience and service effectiveness. Moreover, it can significantly influence  several important applications, such as online advertising \cite{li2015click, Ref3}, information retrieval \cite{wu2014learning}, online product marketing \cite{aggrawal2017brand}, and content recommendation \cite{gonccalves2010popularity}. 

Popularity prediction on social media is usually defined as the problem of estimating the rating scores, view counts, or click-through of a post \cite{Ref19}. In this study, image popularity prediction on social media websites is analyzed to better understand the popularity factors for a particular image. Although this problem has recently received significant attention \cite{Ref11, Ref12, Ref21, Ref24}, it remains a challenging task. For example, image popularity prediction can be significantly influenced by various factors (and features), such as visual content, aesthetic quality, user, post metadata, and time; therefore, considering all this multimodal information is crucial for an efficient prediction. Moreover, it is nontrivial to select an appropriate model that can make better use of the various features contributing to image popularity and accurately predict it. For example, simple machine learning schemes (e.g., support vector and decision tree regression) learn to predict by being fed with highly structured data, thus requiring time and skill to fine-tune the hyperparameters. However, to obtain accurate prediction results, it is critical to construct a prediction model capable of learning through a more abstractive data representation and optimizing the extracted features.

Accordingly, we address the image popularity prediction problem by analyzing a large-scale dataset collected from Flickr to investigate two essential components that may contribute to the popularity of an image; namely, visual content and social context. In particular, we examine the effect of the visual content of an image on its popularity by adopting different types of features that describe various visual aspects of the image, including high-level, low-level, and deep learning features. These are extracted by applying several techniques from machine learning and computer vision. Additionally, we explore the significant role of social context information associated with images and their owners by analyzing the following three types of social features: user, post metadata, and time. To demonstrate the efficacy of the proposed features, we propose a computational deep learning model, called visual-social convolutional neural network (VSCNN), which uses two individual CNNs to learn high-level representations of the visual and social features independently. The outputs of the two networks are then merged into a shared network to learn joint multimodal features and compute the popularity score in the output layer. End-to-end learning is employed to train the entire model, and the weights of its parameters are learned through back-propagation. In summary, the main contributions of this study are as follows: 

\begin{itemize}
\itemsep0em 
\item We demonstrate a comprehensive exploration of the independent benefits and predictive power of various types of visual and social context features towards the popularity of an image. We further demonstrate that these multimodal features can be combined effectively to enhance prediction performance.   
\item A deep learning VSCNN model is proposed for predicting image popularity on social media. VSCNN uses dedicated CNNs to learn structural and discriminative representations from the input visual and social features, achieving considerable performance in predicting image popularity compared with several other traditional machine learning schemes.
%\item We effectively set the architectures and parameters of the adopted CNNs to fit the multimodal information, i.e., social information and visual content of the image.
\item We demonstrate that processing visual and social features using the late fusion scheme is significantly better than using the early fusion scheme.
\item A large-scale dataset of approximately 432K images posted on Flickr is used to evaluate the performance of the proposed VSCNN model. The simulation results demonstrate that VSCNN achieves competitive performance and outperforms six baseline models and other state-of-the-art methods.
%, with a relative improvement of more than 2.33\%, 7.59\%, and 14.16\% in %terms of Spearman's Rho, mean absolute error, and mean squared error, %respectively.

\end{itemize}

The rest of the paper is organized as follows. Related work is reviewed in Section \ref{sec:Related_works}. The extraction of visual and social features is then described in Section \ref{sec:features}. Section \ref{sec:VSCNN} presents the details of the proposed framework for image popularity prediction and the six baseline models used for comparison. The experimental setup and results are discussed in Section \ref{sec:Experiments}. Section \ref{sec:conc} concludes the study and provides some directions for future work. 

%The remainder

\section{related work}
\label{sec:Related_works}
In recent years, predicting the popularity of social media content has received substantial attention \cite{Ref24,Ref5,Ref39,Ref18}. Regarding image popularity prediction,  the related studies differ in terms of the definition of the popularity metric (e.g., view, reshare, and comment counts); however, they all share the same basic pipeline consisting of extracting and testing several types of features that influence popularity, followed by applying a classification or regression model for prediction. Therefore, we review these studies by categorizing them according to the features and prediction models used.

%\subsection{Used Features} 
%\label{sec:feature used}

Regarding the features used, the existing studies have primarily focused on investigating the relative effectiveness of various feature types for predicting image popularity, including social context, visual content, aesthetic, and time. For instance, Khosla \textit{et al.} \cite{Ref13} demonstrated that image content (e.g., gist, color histogram, texture, color patches, gradient, and deep learning features) and social cues (e.g., number of followers or number of posted images) have a significant effect on image popularity. Gelli \textit{et al.} \cite{Ref12} employed visual sentiment features along with context and user features to predict a succinct popularity score of social media images. They demonstrated that sentiment features are correlated with popularity and have considerable predictive power if they are used together with context features. Cappallo \textit{et al.} \cite{Ref11} demonstrated that latent image features can be used to predict image popularity. They explored the visual cues that determine popularity by identifying themes from both popular and unpopular images. McParlane \textit{et al.} \cite{Ref21} performed image classification using a combination of four broad feature types, that is, image content, image context, user context, and tags, to predict whether an image will obtain a high or low number of views and comments in the future.
 
Compared to the aforementioned approaches, relatively few studies have been conducted to demonstrate the effect of time and aesthetic features on image popularity. For instance, Wu \textit{et al.} \cite{Ref19} developed a new framework called multi-scale temporal decomposition to predict image popularity based on popularity matrix factorization. They explored the mechanism of dynamic popularity by factoring popularity into two contextual associations, i.e., user-item context and time-sensitive context. Furthermore, Almgren \textit{et al.} \cite{almgren2016predicting} employed social context, image semantics, and early popularity features to predict the future popularity of an image. Specifically, they considered the popularity changes over time by collecting information regarding the image within an hour of uploading and keeping track of its popularity for a month. Totti \textit{et al.} \cite{totti2014impact} analyzed the effect of visual content on image popularity and its propagation on online social networks. Along with social features, they proposed using aesthetic properties and semantic content to predict the popularity of images on Pinterest.

We observe that most of the aforementioned studies rely only on a part of the useful features for image popularity prediction, and do not consider the interactions between other pertinent types of features.

%\subsection{Prediction Models} 
%\label{sec:models}
Regarding the models used for prediction, previous studies have introduced several types of machine learning schemes. Both \cite{Ref13} and \cite{Ref12} considered image popularity prediction as a regression problem in which support vector regression (SVR) \cite{Ref57} was used to predict the number of views that an image received on Flickr. Totti \textit{et al.} \cite{totti2014impact} reduced the problem to a binary classification task and utilized a random forest classifier \cite{liaw2002classification} to predict whether an image would be extremely popular or unpopular based on the number of reshares on Pinterest. Moreover, the authors in \cite{wang2017combining} predicted the number of views for an image on Flickr using a gradient boosting regression tree \cite{Ref54}. Although most of these prediction models perform satisfactorily, they tend to generate smoothed results, making the popularity of images with overly high or low scores difficult to predict accurately. In addition, it may be time-consuming to fine-tune the hyperparameters that significantly influence the performance of these models.

Recently, deep learning techniques have gained widespread attention and achieved outstanding performances in various tasks \cite{simonyan2014very, szegedy2015going,abousaleh2016novel,krizhevsky2012imagenet}, owing to the capability of deep neural networks to learn complex representations from data at each layer, where they imitate learning in the human brain by abstraction. Nevertheless, insignificant effort has been expended for predicting image popularity using these techniques. In this regard, Wu \textit{et al.} \cite{wu2017sequential} proposed a new deep learning framework to investigate the sequential prediction of image popularity by integrating temporal context and attention at different time scales. Moreover, Meghawat \textit{et al.} \cite{meghawat2018multimodal} developed an approach that integrates multiple multimodal information into a CNN model for predicting the popularity of images on Flickr. Although these studies have achieved satisfactory performances, they are not sufficiently powerful to capture and model the characteristics of image popularity. For instance, the authors in \cite{meghawat2018multimodal} investigated the effect of the visual content of an image on its popularity by utilizing only one feature obtained by the pre-trained InceptionResNetV2 model, whereas they ignored other important visual cues, such as low-level computer vision, aesthetics, and semantic features. Moreover, although it has been demonstrated that time features have a crucial effect on image popularity \cite{Ref19, wu2017sequential}, they were not considered in the proposed model. They also adopted an early fusion scheme for processing the proposed multimodal features, despite several studies having demonstrated that this scheme is outperformed by the late fusion scheme in processing heterogeneous information \cite{ Ref13,hou2018audio}.

To address the aforementioned issues, we propose a multimodal deep learning prediction model that uses numerous types of features associated with image popularity, including multi-level visual, deep learning, social context, and time features. The proposed model uses dedicated CNNs for separately learning high-level representations from the input features and then efficiently merges them into a unified network for popularity prediction.

%In this paper, we initially analyze a real dataset gathered from Flickr to identify and extract different kinds of features that are associated with the popularity of the image, including multi-level visual features and social context information. Based on these features and the recent success of convolutional neural networks (CNNs) in many fields, we propose a visual-social CNN (VSCNN) model for image popularity prediction. In the proposed model, visual and social features are first separately processed utilizing two individual CNNs and then merged into a united network to compute the popularity score.

\section{features}
\label{sec:features}
In this section, various types of features that can influence image popularity are analyzed. First, in Section \ref{sec:A3}, we investigate the visual features that could be used to describe the different facets of images based on their content, including low-level, high-level, and deep learning features. Then, in Section \ref{sec:A7}, several social features based on the contextual information of images and their owners are explored. 

\subsection{Visual Content Features} 
\label{sec:A3}

\subsubsection{Low-level Features}
\label{sec:A4}
%\paragraph{Low-level features}

There are several types of low-level computer vision features (e.g., texture, color, shape, gist, and gradient) that are likely to be used for visual processing. In this study, the three following features are adopted: color, texture, and gist.  

\textbf{Color:} A perfect color distribution in an image attracts viewer attention and aids in determining object properties and understanding scenes. In this study, a color histogram descriptor that results in a vector of 32 dimensions and characterizes the color feature is used \cite{hassanien2008computational}.

\textbf{Texture:} Texture can define the homogeneity of colors or intensities of an image. It can also be utilized to identify the most interesting objects or regions \cite{Ref29}. To investigate its effect on image popularity, we employ one of the most widely used features for texture description, namely,  local binary patterns (LBP) \cite{Ref28}. More precisely, the uniform LBP descriptor is used \cite{Ref30}, resulting in a 59-dimensional feature vector.

\textbf{Gist:} The GIST descriptor provides a rough description of a scene by epitomizing the gradient information (scales and orientations) for various parts of a photo. To extract the GIST feature of an image, we adopt the widely used GIST descriptor proposed in \cite{Ref31}, resulting in a feature vector with 512 dimensions.

%The GIST descriptor offers a rough description of a scene by epitomizing the gradient information (scales and orientations) for various parts of a photo. In this work, we adopt the widely used GIST descriptor proposed by \cite{Ref31}, to generate a feature vector with 512 dimensions.

%Basically, people can recognize scenes only by looking at their gist. Therefore, the GIST descriptors are developed to match the human conceptions relative to different features of photos.

%Basically, people can recognize scenes only by looking at their gist. The GIST descriptors are therefore designed to match human concepts with respect to various features of images.  The GIST descriptor offers a rough description of the scene by epitomizing the gradient information (scales and orientations) for various parts of the photo. To extract the GIST feature of the image, we chose to use the popular GIST descriptor proposed by \cite{Ref31} with a feature dimension of 512.

% We adopt the widely used Gist descriptor proposed by \cite{Ref31} (to generate a feature vector with 512 dimensions.)with a feature dimension of 512. 

\subsubsection{High-level Features}
\label{sec:A5}
The quality and aesthetic appearance of an image are important for its popularity. Based on the various photographic techniques and the aesthetic standards used by professional photographers, we adopt certain aesthetic features for popularity prediction. These features are developed to evaluate the visual quality of a photograph by separating the subject area from the background using the blur detection technique \cite{Ref33}. Then, based on the result of this separation process, six types of aesthetic features are computed as indicated below.

\begin{itemize}
\item \textbf{Clarity contrast:} To gain viewer attention to the key point of a photograph and to isolate the subject region from the background, professional photographers normally adjust the lens to keep the subject in focus and make the background out of focus. Accordingly, a clear photograph will comparatively have more high-frequency components than a blurred photograph. To characterize this property, we define a clarity contrast feature based on the method presented in \cite{Ref33}. 
 
%\begin{equation}
%\label{eq:11}
%\textit f_{c}= \frac{ \frac{\norm{M_S}}{ \norm{S}}}{ \frac{\norm{M_I}}{ \norm{I}}},
%\end{equation}
%where $\norm{S}$ and $\norm{I}$ are the areas of the subject region and the original photograph, respectively, and
%\begin{equation}
%\label{eq:12}
%\textit M_I = \big\{(m,n)\mid \abs{F_I(m,n)}> \gamma\mbox{ max} \big\{F_I(m,n)\big\} \big\} ,
%\end{equation}
%\begin{equation}
%\label{eq:13}
% %\hspace{2mm}
%\textit M_S = \big\{(m,n)\mid \abs{F_S(m,n)}> \gamma\mbox{ max} \big\{F_S(m,n)\big\} \big\},
%\end{equation}
%\begin{equation}
%\label{eq:14}
%\textit F_I = FFT(I),\mbox{ } F_S = FFT(S).
%\end{equation}
%
%$\frac{\norm{M_S}}{ \norm{S}}$ and $\frac{\norm{M_I}}{ \norm{I}}$ are the ratios of the area of the high-frequency components to the area of all frequency components in $S$ and $I$, respectively. $m$ and $n$ are the spatial frequencies. $\gamma$ is a predefined threshold set to 0.2 in our experiments. $FFT(I)$ and $FFT(S)$ are the fast Fourier transforms calculated over $I$ and $S$, respectively.  

\item \textbf{Hue count:} The hue count of an image is a metric of its simplicity. It can also be used to assess the image quality. Although professional photographs appear bright and vivid, their hue number is normally less than that of amateur photographs. Therefore, we compute the hue count feature of an image using a 20-bin histogram $H_c$, which is computed on the good hue values. This can be formulated as follows \cite{ke2006design}: 
 
\begin{equation}
\label{eq:15}
\textit f_l = 20 - \abs{N_c},
\end{equation}
\begin{equation}
\label{eq:16}
N_c = \big\{i \mid H_c(i)>\beta m\big\}.
\end{equation}
where $N_c$ denotes the set of bins with values larger than $\beta m$, $m$ is the maximum histogram value, and $\beta$ is used to control the noise sensitivity of the hue count. We selected $\beta=0.05$ in our experiments.

\item \textbf{Brightness contrast:} In high-quality photographs, the subject area's brightness significantly differs from that of the background because professional photographers frequently use different lighting on the subject and the background. Nevertheless, most amateurs use natural lighting and allow the camera to adjust the brightness of a picture automatically; this usually reduces the difference in brightness between the subject area and the background. To discern the difference between these two types of photographs, we calculated a brightness contrast feature based on the method described in\cite{tang2013content}.

%The brightness difference between two neighboring surfaces is called the brightness contrast. In high-quality photographs, the subject area's brightness significantly differs from that of the background, because professional photographers frequently utilize different subject and background lightings. Nevertheless, most of the amateurs use natural lighting and allow the camera to automatically adjust the brightness of a picture, which usually reduces the difference in brightness between the subject area and the background. To discern the difference between these two types of photographs, the brightness contrast feature $f_b$ is calculated as \cite{tang2013content}:

%\begin{equation}
%\label{eq:17}
%f_b = \abs{\textrm{ln}\Big(\frac{B_s}{B_b}\Big)},
%\end{equation}
%where $B_s$ and $B_b$ denote the average brightness of the subject region and the background, respectively.

\item \textbf{Color entropy:} Owing to the distinct interrelationship between the color planes of drawings and natural images, the entropy of RGB and Lab color space components is computed to differentiate natural images from drawings \cite{chen2019no}.

\item \textbf{Composition geometry:} The proper geometrical composition is a fundamental requirement for obtaining high-quality photographs. The rule of thirds is one of the most important photographic composition principles utilized by professional photographers to bring more balance and high quality to their photos. To formulate this criterion, we define a composition feature based on the method introduced in \cite{tang2013content}.

%It states that putting the significant subjects of images along the imagery thirds lines or around their intersections often produces highly aesthetic photos.
% 

%The proper geometrical composition is a fundamental demand to get high-quality photographs. The \textit{Rule of Thirds} is one of the effective tools of photographic composition. It commonly utilized by professional photographers to bring more balance and interest in their photos. The \textit{Rule of Thirds} works by imagining that the photo is split into nine quadrants of equal size using two equally spaced vertical lines and two equally spaced horizontal lines. Then, the subjects of the photo and any important compositional objects are put along those lines or their intersections to create an image with a better visual impact. The composition geometry feature is computed as \cite{tang2013content}: 
%
%\begin{equation}
%\label{eq:20}
%f_m = \min_{j=1,2,3,4}\sqrt{\frac{\big(C_{s_{x}}-P_{j_{x}}\big)^2}{W^2} +\frac{\big(C_{s_{y}}-P_{j_{y}}\big)^2}{H^2}}.
%\end{equation}
%
%where $\big(C_{s_{x}},C_{s_{y}}\big)$ is the centroid of the subject area, $\big(P_{j_{x}},P_{j_{y}}\big)$, $j= 1,2,3,4,$ are the four intersection points in the photo, and $W$ and $H$ represent the photo's width and height, respectively.

\item \textbf{Background simplicity:} Professional photographers normally maintain simplicity within the shot to enhance the composition of the photo. This is because photographs that are clean and free of distracting backgrounds are considerably more appealing and naturally draw the attention of a viewer to the subject. The color distribution in a simple background tends to be less dispersed. Therefore, we compute a feature that represents the background simplicity of an image using the method proposed in \cite{Ref33}, which is based on the color distribution of the background.

\end{itemize}

In this study, we combine the six aesthetic features indicated above, resulting in an 11-dimensional feature vector.

\subsubsection{Deep Learning Features}
\label{sec:A6}
\begin{figure}[htbp]
\begin{center}
\centerline{\includegraphics [width=93mm,scale=1,height=6cm]{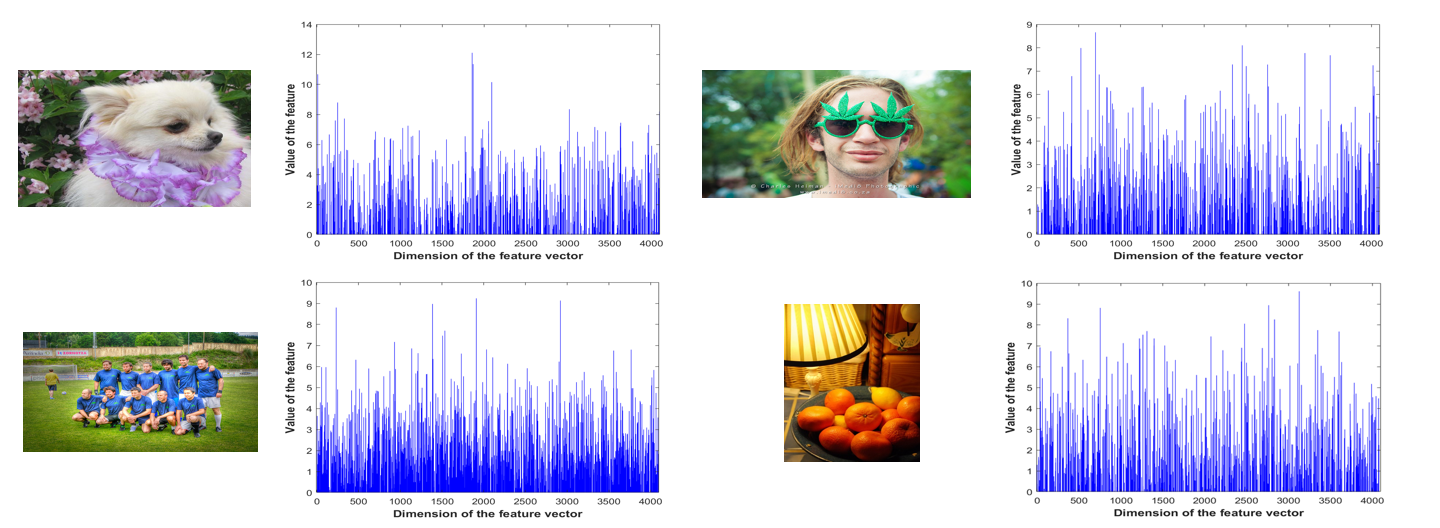}}
\captionsetup{}
\caption{The plots of deep learning feature vector values of different images from the dataset.}
\label{fig4}
\end{center}
%\vspace{4mm}
\end{figure}

Recently, deep learning methods have been widely used for image representation owing to their effectiveness \cite{simonyan2014very}, \cite{krizhevsky2012imagenet}. In this study, the CNN architecture of the VGG19 model was employed to learn the deep features of photographs \cite{simonyan2014very}. The VGG19 model was trained on 1.2 million images from the ImageNet database to classify these images into 1000 categories  \cite{krizhevsky2012imagenet}. The Keras framework of the VGG19 pre-trained CNN model \cite{Ref16} was used for feature extraction from the layer situated immediately prior to the final classification layer, (i.e., the last fully connected layer (fc7)). The output of this layer is a 4,096-dimensional feature vector. A few images selected from the dataset and the plots of their respective deep feature vector values are shown in Fig.~\ref{fig4}.

%.............Figure1......%
\captionsetup[figure]{labelformat={default},labelsep=period,name={Fig.}}
%%%%%%%%%%%%%%%%%%%%% 

\vspace{-3mm}

\subsection{Social Context Features} 
\label{sec:A7}
Previous studies demonstrated that the popularity of an image depends not only on its content, but also the social information regarding the user uploading the image, as well as the textual information associated with it \cite{Ref13,Ref24}. In this section, we attempt to determine the extent to which social features can influence image popularity. The following three types of social features are analyzed: user, post metadata, and time.
\subsubsection{User}
\label{sec:A8}
The popularity of a user highly correlates with the popularity of his/her posted images. Therefore, we adopted several user-centered features, which are listed below, and will have the same value for all photographs posted by the same user.

\textbf{User id:} This is defined as a unique integer number (ranging from 1 to 135) according to the average view count of all images of a user (i.e., a greater average number of views implies a larger value). Thus, this number is a unique identification of each user and can be used directly in the prediction model.

\textbf{Average views:}
%The mean number of views of all the public photos uploaded by the user.
%The mean number of views of all user-uploaded photos.
The average view count of all user-uploaded photos.

\textbf{Group count:} The number of groups to which a user subscribes.
%The number of groups a given user subscribes to.
%The number of groups to which a given user subscribes.
%The number of groups to which the given user is a member.

\textbf{Member count:} The mean number of members in the groups to which a user subscribes.

%The average number of members in the groups a given user belongs to.
%The mean number of participants in the groups to which a given user belongs.
%The mean number of members in the groups to which the given user is a member.

\textbf{Image count:}
The total number of photographs posted by a user.

To characterize the effects of these features on predicting image popularity, we computed their rank correlation with the image popularity score using Spearman’s rank correlation coefficient (Spearman’s Rho) \cite{Ref43}. The value of the correlation coefficient ranges from [-1, 1], where a score of 1 (resp. -1) indicates an ideal positive (resp. negative) association, and a score of zero indicates no correlation. The results are shown in Table~\ref{tab1}; both, user id and image views have a strong positive correlation with image popularity (Spearman’s Rho = 0.74).

%%.....Table1...............%%
\begin{table}
\renewcommand{\arraystretch}{1.3}
\caption{Spearman's Rho values for the correlation of user features with popularity score.}
\label{tab1}
\renewcommand{\arraystretch}{1.3}
\setlength{\tabcolsep}{3pt}
\centering
\begin{tabular}{|P{60pt}|P{60pt}|}
    \hline
    \textbf{Feature}  &  \textbf{Spearman's Rho}\\
    \hline
    User id         &   \textbf{0.74}\\
    \hline

    Average views   &  \textbf{0.74}\\
    \hline
    Group count     &   0.067\\
    \hline
    Member count    &   0.19\\
    \hline
    Image count     &   - 0.35\\
     \hline
\end{tabular}
\newline
\vspace{-3mm}
\end{table}

\subsubsection{Post Metadata}
\label{sec:A9}
The contextual information associated with an uploaded image (e.g., tags, comments, or title) can also influence its popularity. For instance, an image with a large number of tags is expected to appear more frequently in search results. Therefore, we consider certain image contextual features for popularity prediction, which refer to image-related metadata, and most are entered by the user. The image-context features adopted in this study are listed below.

\textbf{Tag count:} The number of tags annotated by a user on a posted photograph.

\textbf{Title length:} The number of characters in the title of a photograph.
 
\textbf{Description length:} The character count in the image description.
%The image description length.
%The character count in the image description.

\textbf{Tagged people:} A binary number (1 or 0) indicating whether a given photograph has tagged people or not.

\textbf{Comment count:} The number of comments an image has obtained from other users.

%%.....Table2...............%%
\begin{table}
\renewcommand{\arraystretch}{1.3}
\caption{Spearman's Rho values for the correlation of post metadata features with popularity score.}
\label{tab2}
\renewcommand{\arraystretch}{1.3}
\setlength{\tabcolsep}{3pt}
\centering
\begin{tabular}{|P{60pt}|P{60pt}|}
    \hline
    \textbf{Feature}  &  \textbf{Spearman's Rho}\\
    \hline
    
    Tag count       &   0.49\\
    \hline

    Title length   &  0.22\\
    \hline
     Description length    &  0.51\\
    \hline
     Tagged people    &   0.0043\\
     \hline
\end{tabular}
\newline
\vspace{-4mm}
\end{table}
%%%%%%%%%%%%%%%%%%%%%%%%%%%%

We calculated the relationship between each of these features and the popularity of an image, as conducted for user features. The results of Spearman's Rho are listed in  Table~\ref{tab2}. Note, most of the post features have a significant positive correlation with popularity, except the tagged people feature, which has a slight positive correlation of 0.0043.

Considering the results shown in Tables~\ref{tab1} and~\ref{tab2}, the user id and image views features have the highest Spearman’s Rho scores, which implies that user-centric features are the most effective in predicting image popularity. This also agrees with what we expected because popular users usually have a significant number of followers, indicating their images are more likely to receive a larger number of views after uploading on social media and thus become popular.
\subsubsection{Time}
\label{sec:A10}
Along with the user and post features for predicting image popularity, there is a strong dependence on time features. For instance, users tend to become more active on social websites during the weekend. Thus, images posted at these time slots would naturally be expected to receive more views and therefore more ratings. Hence, we consider the following time features: post day, post month, post time, and post duration. The definitions of these features are as follows:

\textbf{Post day:}
The day of the week on which a photograph is posted. We encoded the day number using one-hot encoding of a 7-dimensional vector.

\textbf{Post month:}
The month in which a photograph is posted. We encoded the month number using one-hot encoding of a 12-dimensional vector.

\textbf{Post time:}
The period of day during which a photograph is posted. The day was divided into four segments (six hours in each segment) assuming that the photograph is posted either in the morning (06:00 to 11:59), afternoon (12:00 to 17:59), evening (18:00 to 23:59), or night (00:00 to 05:59). Then, the post time was encoded using the one-hot encoding of a 4-dimensional vector.

\textbf{Post duration:}
%The time duration in days since the uploaded day of the image on Flickr.
The amount of time in days during which the photograph remained posted on Flickr.

%In summary, we adopt both visual and social features to predict the popularity of the image. Consequently, we concatenate all the extracted features above to obtain a 4,744-dimensional feature vector which can be used for popularity prediction. 

\section{methodology}
\label{sec:VSCNN}
In this section, we explain the details of the proposed framework for predicting social media image popularity; moreover, we present a brief description of the baseline models.

\subsection{Overview of Proposed Framework}
\label{sec:Overview }
\begin{figure*}[htbp]
\begin{center}
\centerline{\includegraphics [width=\textwidth, height=8cm]{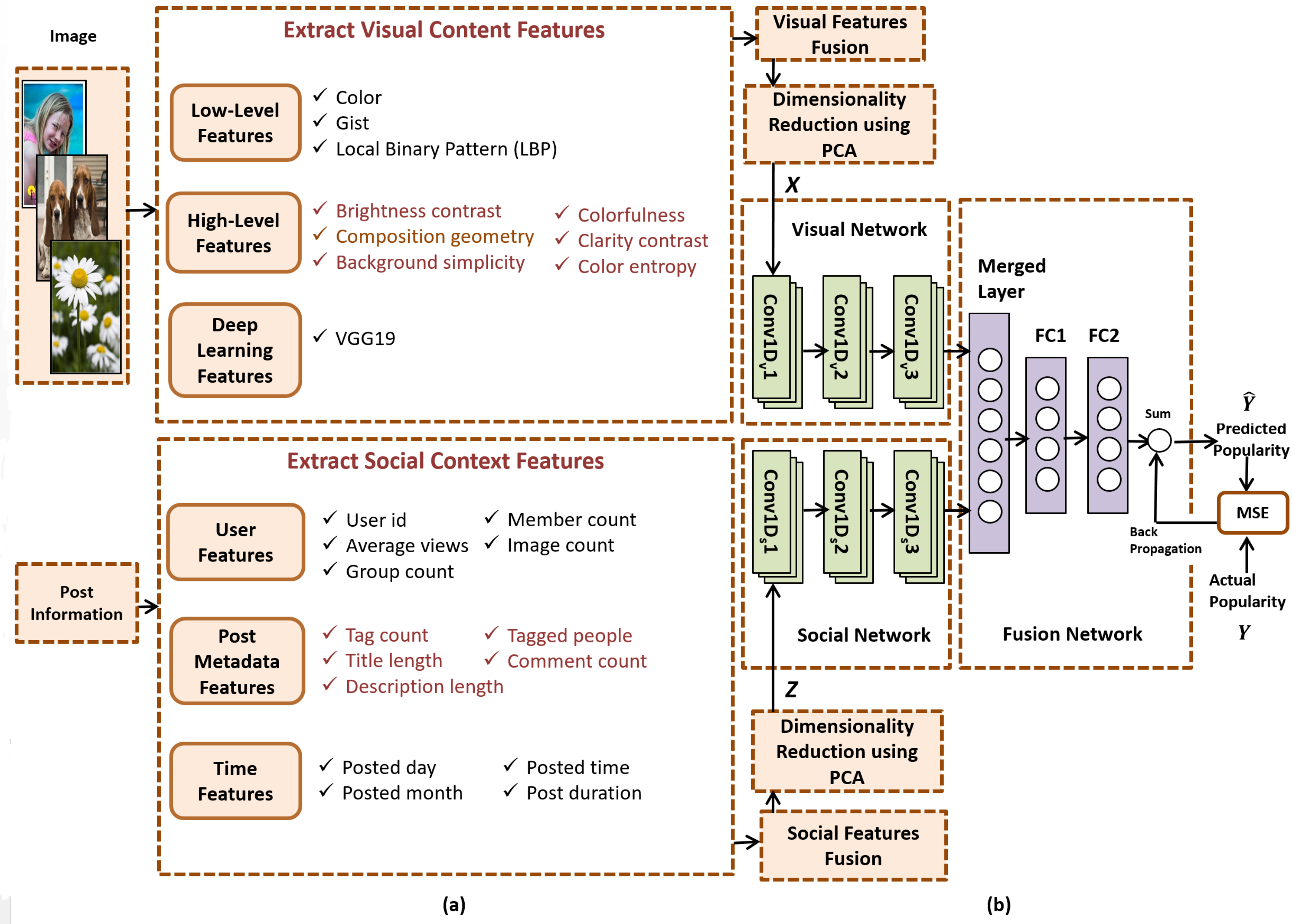}}
\caption{Diagram of the proposed framework for image popularity prediction. (a) Feature extraction, and (b) Proposed VSCNN regression model.}
\label{framework}
\end{center}
\vspace{2mm}
\end{figure*}

The overall diagram of the proposed framework is shown in Fig.~\ref{framework}. It consists of the following two phases: feature extraction and VSCNN regression model. In the feature extraction phase, for each post in the dataset, we extract visual features from the image and social context features from its corresponding post-context information, as shown in Fig.~\ref{framework} (a). The extracted visual features are integrated to obtain a final feature vector of 4,710 dimensions that describe the different visual facets of the image. Then, principal component analysis (PCA) \cite{jolliffe2016principal} is performed to decrease the dimensionality of this vector from 4,710 to 20 and to select only the prevalent features. This results in a visual-PCA descriptor of 20 dimensions, which is denoted by $X$. Finally, the values of the  $X$ features are normalized so that all of them belong to the same scale. Similarly, the same procedure is applied to the corresponding extracted social features to obtain a normalized social-PCA descriptor of 14 dimensions, which is denoted by $Z$. The obtained $X$ and $Z$ are used as inputs to the proposed VSCNN model to predict the popularity of the corresponding post. 

As shown in Fig.~\ref{framework} (b), the proposed VSCNN model consists of two individual CNNs that are used to derive the structural and discriminative representations from the visual ($X$) and social ($Z$) features; namely the visual network and social network. Each of these networks is a one-dimensional CNN consisting of three convolutional layers. The rectified linear unit (ReLU) is employed as the activation function for each convolutional layer to avoid the vanishing gradient problem in the training phase. A fusion network is used to combine the outputs of these networks into a unified network. It consists of one merged layer and two fully connected layers. The merged layer is used to concatenate the outputs of the last convolutional layer of the visual network and that of the social network and generate the inputs of the first fully connected layer. Finally, the outputs of the second fully connected layer are summed at the final node, generating the predicted popularity score. In the diagram, the convolutional and fully-connected layers are denoted by ${\mathrm{Conv1D}}_{\mathrm{v}}1$, ${\mathrm{Conv1D}}_{\mathrm{v}}2$, ${\mathrm{Conv1D}}_{\mathrm{v}}3$, ${\mathrm{Conv1D}}_{\mathrm{s}}1$, ${\mathrm{Conv1D}}_{\mathrm{s}}2$, ${\mathrm{Conv1D}}_{\mathrm{s}}3$, FC1, and FC2, where the subscripts “v” and “s” indicate visual and social features, respectively.

%${\mathrm{Conv1D}}_{\mathrm{v}}1$, ${\mathrm{Conv1D}}_{\mathrm{v}}2$, ${\mathrm{Conv1D}}_{\mathrm{v}}3$, ${\mathrm{Conv1D}}_{\mathrm{s}}1$, ..., FC1, and FC2, where the subscripts 'v' and 's' denote the visual and social feature, respectively.

%.............Figure2......%
\captionsetup[figure]{labelformat={default},labelsep=period,name={Fig.}}

\subsection{Training the VSCNN Model}
\label{sec:Training }
Similar to the other CNNs, we first prepare a set of training samples ($N$) to train the VSCNN model. Each sample comprises of the posted image, post-context information, and corresponding popularity score ($Y$). We then calculate the visual feature descriptor ($X$) and the corresponding social feature descriptor ($Z$) for each sample, as indicated above. For each iteration, we obtain the output of the visual network as 
%\useshortskip
\begin{equation}
\label{eq:21}
V_i= Conv1D_{v}3\Big(Conv1D_{v}2\big(Conv1D_{v}1(X_i)\big)\Big), \mbox{ } i= 1... N.
\end{equation}
Similarly, the output of the social network is as follows:
\begin{equation}
\label{eq:22}
S_i= Conv1D_{s}3\Big(Conv1D_{s}2\big(Conv1D_{s}1(Z_i)\big)\Big), \mbox{ } i= 1... N.
\end{equation}
Next, we flatten $V_i$ and $S_i$, and concatenate the two feature vectors using a merge layer, and then we use the  resulting concatenated feature vector as the input of the fusion network, $F_i = [V_{i}^{'} \mbox{ }  S_{i}^{'}]^{'}$. Thus, a fully connected cascade-feed-forward network can be calculated as follows:

\begin{equation}
\label{eq:23}
\hat{Y_i}= FC2\Big(FC1\big(F_i)\Big), \mbox{ } i= 1... N.
\end{equation}
Let $\theta$ denote the parameters of the VSCNN model. First, they are initialized using random values ranging between -1 and 1, and then are trained by optimizing the following mean squared error (MSE) cost function using back-propagation:
\useshortskip
\begin{equation}
\label{eq:24}
MSE(\theta)= \min_{\theta}\Big(\frac{1}{N}\sum_{i=1}^{N}\left\|\hat{Y_i}-Y_i\right\|_{2}^{2}\Big).
 \end{equation}
%\norm{\hat{Y_i}-Y_i}_{2}^{2}
A stride of size 1 was adopted in the networks of the VSCNN model. To avoid overfitting, a dropout of 0.1 was adopted after each layer, except for the last fully connected layer, in which a dropout of 0.2 was used. Additionally, batch normalization was applied to each convolutional layer to increase the stability of the CNNs. Further details regarding the configuration of the VSCNN model are presented in Table~\ref{tab3}.

%%% Table3%%%%%
\begin{table}
\caption{Configuration of the VSCNN Model. }
\label{tab3}%\setlength{\tabcolsep}{3pt}
\centering
\begin{tabular}{|P{50pt}|P{30pt}|P{50pt}|P{50pt}|}
%\begin{tabular}{c|c|c|c}
\hline
Layer& 
Kernel& 
Activation Function &
Number\par of Neurons
 \\
\hline
${\mathrm{Conv1D}}_{\mathrm{v}}1$& 3&ReLU&32\\
\hline
${\mathrm{Conv1D}}_{\mathrm{v}}2$&3&ReLU&64\\
\hline
${\mathrm{Conv1D}}_{\mathrm{v}}3$&3&ReLU&128\\
\hline
${\mathrm{Conv1D}}_{\mathrm{S}}1$& 2&ReLU&32\\
\hline
${\mathrm{Conv1D}}_{\mathrm{S}}2$& 2&ReLU&64\\
\hline
${\mathrm{Conv1D}}_{\mathrm{S}}3$& 2&ReLU&128\\
\hline
Merged Layer& & & 4736\\
\hline
FC1& &ReLU&1024 \\
\hline
FC2&&ReLU& 500\\
\hline
\end{tabular}
\newline
\vspace{-3mm}
\end{table}

%.............Figure3......%
\captionsetup[figure]{labelformat={default},labelsep=period,name={Fig.},skip=0pt}
\begin{figure}[htbp]
\begin{center}
\centerline{\includegraphics [width=85mm,scale=1]{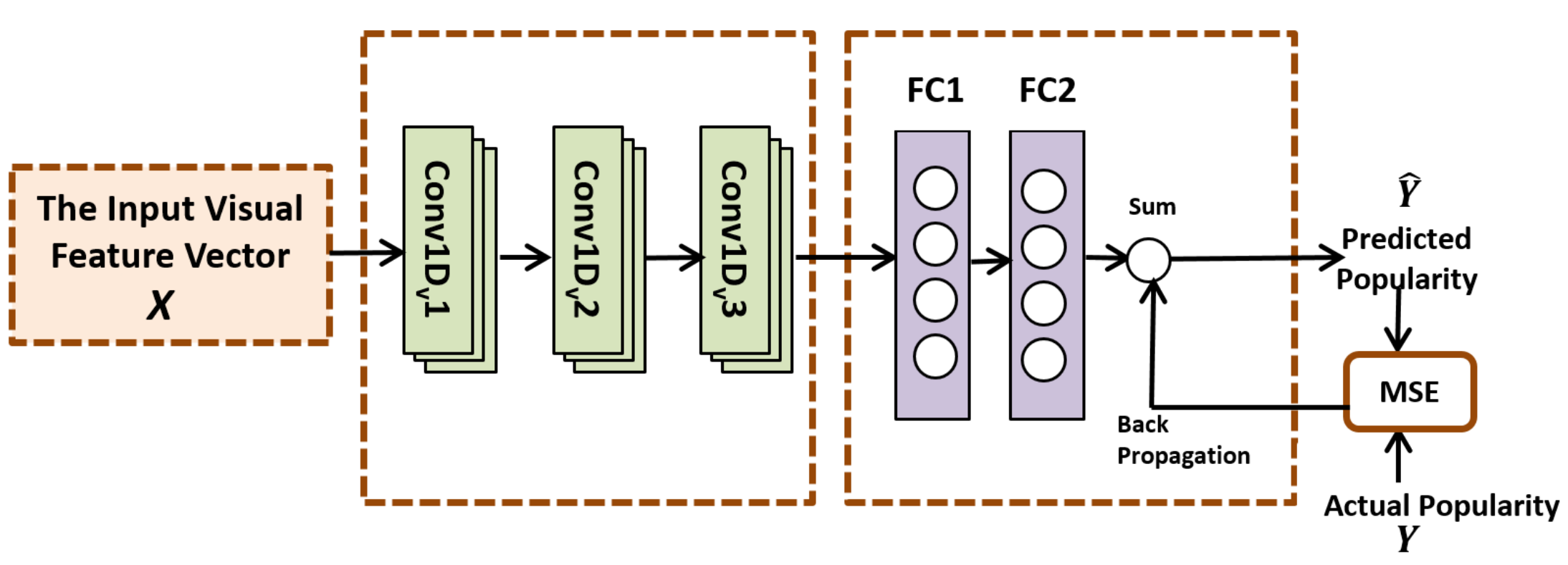}}
\caption{Structure of the VCNN model.}
\label{VCNN}
\end{center}
\vspace{-2mm}
\end{figure}
%%%%%%%%%%%%%%%%%%%%% 

%.............Figure4......%
\captionsetup[figure]{labelformat={default},labelsep=period,name={Fig.},skip=0pt}
\begin{figure}[htbp]
\begin{center}
\centerline{\includegraphics [width=85mm,scale=1]{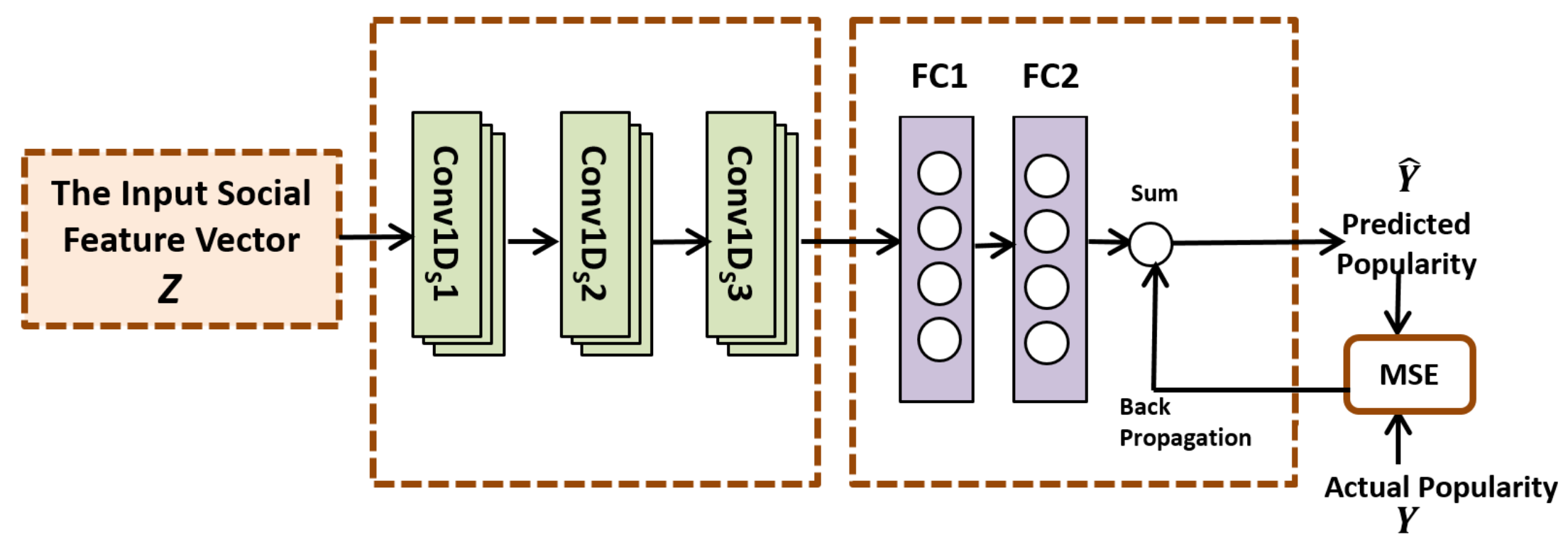}}
\caption{Structure of the SCNN model.}
\label{SCNN}
\end{center}
\vspace{-4.5mm}
\end{figure}
%%%%%%%%%%%%%%%%%%%%% 

\subsection{Baseline Models}
\label{sec:Baseline }
Considering that the popularity prediction for social media images is a regression problem, only a small number of current machine learning models can be directly used. Therefore, we first compare the proposed VSCNN model with two CNN-based models; namely the visual-only CNN (VCNN) and the social-only CNN (SCNN) model. Then, we compare the proposed VSCNN with the following four conventional regression models: linear regression (LR), SVR, decision tree regression (DTR), and gradient boosting decision tree (GBDT). A brief description of each model is presented in the following subsections.

%We provide a short explanation (brief summary) of each of these models in the following subsections.
\subsubsection{Visual-only Convolutional Neural Network }
\label{sec:VCNN}
As shown in Fig.~\ref{VCNN}, the VCNN model disjoints all social-related parts in the proposed VSCNN (cf. Fig.~\ref{framework} (b)) and retains the remainder. The VCNN is trained using the same procedure as in the VSCNN model, which was presented in Section \ref{sec:Training }.

\subsubsection{Social-only Convolutional Neural Network}
\label{sec:SCNN}
The structure of the SCNN model is shown in Fig.~\ref{SCNN}. It detaches all visual-related components in the proposed VSCNN model (cf. Fig.~\ref{framework} (b)).

\subsubsection{Linear Regression }
\label{sec:LR}
%Linear regression is a statistical model used to model the relationship between a single dependent output variable ($\textit{y}$) and one or more independent input variables ($\vect{x}$) by finding out a linear function that best describes the input variables. 

%LR is a statistical model utilized for modeling the relationship between a single output variable (dependent) and one or set of input variables (independent )  by finding a linear regression function that best describes the input variables and then can be used to make predictions about tested data.

LR is a statistical model designed for modeling the relationship between a single dependent variable (output) and a set of independent variables (inputs) by finding a linear regression function that best describes the input variables. To predict the popularity score of an image using this model, a linear relationship between the features of an input image and the popularity score was assumed as follows: \useshortskip
\begin{equation}
\label{eq:2}
\textit{y}=\textit{w}_{0}+\textit{w}_{1}\textit{x}_{1}+...+\textit{w}_{n}\textit{x}_{n}+\varepsilon= \vect{x}^\mathsf{T}\vect{w}+\varepsilon,
\end{equation}
where $\textit{y}$ indicates the predicted popularity score of the input image, $\vect{x}$ denotes the feature vector, $\vect{w}$ is the model weight vector, and $\varepsilon$ is the error term. The gradient descent algorithm \cite{Ref44} was employed to learn the weight coefficients during the training phase.  
\subsubsection{Support Vector Regression }
\label{sec:SVR}
SVR \cite{Ref57} is a regression version of a support vector machine \cite{Ref58}. It can construct advanced optimal approximation functions using training data. Given $M$ training samples of popularity feature vectors $\{\vect{x}_{1},\vect{x}_{2}, ..., \vect{x}_{M}\}$, where $\vect{x}_{i}\in\mathbb{R}^d$, and their corresponding popularity score values $\{{y}_{1}, {y}_{2}, ..., {y}_{M}$\}, where ${y}_{i}\in\mathbb{R}$, the regression is performed by determining a continuous mapping $f:\mathbb{R}^d \rightarrow \mathbb{R}$ that best predicts the set of training samples with the approximation function $y = f(\vect{x})$. This is defined as follows: \useshortskip

\begin{equation}
\label{eq:8}
y = f(\vect{x})= \sum_{i=1}^{M}(\alpha_{i}-\alpha_{i}^{*})K(\vect{x}_{i},\vect{x})+ b,
\end{equation}
where $\alpha_{i}$ and $\alpha_{i}^{*}$ are the Lagrange multipliers associated with each training sample $\vect{x}_{i}$, $K$ denotes the kernel function, and $b$ is the bias term. In this study, the Gaussian radial basis function (RBF) \cite{Ref60} was utilized as the kernel function.

%During the training process, the SVR algorithm attempts to find an optimal set of $\alpha_{i}$ and $\alpha_{i}^{*}$, $\forall i\in [1,M]$, by maximizing the following dual objective function:
%\useshortskip 
%\begin{equation}
%\label{eq:9}
%       \begin{gathered}     
%-\dfrac{1}{2}\sum_{i=1}^{M}\sum_{j=1}^{M}(\alpha_{i}-\alpha_{i}^{*})(\alpha_{j}-\alpha_{j}^{*})K(\vect{x}_{i},\vect{x}_{j})\\
%-\epsilon \sum_{i=1}^{M}(\alpha_{i}+\alpha_{i}^{*})+ \sum_{i=1}^{M}(\alpha_{i}-\alpha_{i}^{*})y_{i}
%       \end{gathered}
%\end{equation}
%
%\useshortskip 
%\begin{equation}
%\label{eq:10}
%\text{subject to}  \mbox{   } \mbox{   }
%\sum_{i=1}^{M}(\alpha_{i}-\alpha_{i}^{*})= 0 \mbox{ and} \mbox{ }\alpha_{i},\alpha_{i}^{*}\in[0,\textit{C}]
%\end{equation}
%
%where $\epsilon$ is an insensitive tolerance value that determines the degree of accuracy of the approximated function, and $\textit{C}$ is a penalty parameter that controls the trade-off between the algorithm complexity and the training error. In our experiments, we identified the values of numerous significant parameters of SVR, including C = 3, epsilon ($\epsilon$) = 0.1, gamma = auto, and kernel = RBF.

\subsubsection{Decision Tree Regression }
\label{sec:DTR}
A decision tree can be used to predict the value of a continuous dependent variable from a set of continuous predictors by constructing a predictive model with a tree-like structure. In this study, the classification and regression tree (CART) algorithm \cite{Ref45} was used to construct a decision tree. Using this algorithm, we constructed a model that can predict the popularity score by learning simple decision rules derived from the data features. For each feature, the CART algorithm splits the data at different points, then selects the part that minimizes the sum of squared errors (SSE) and generates more homogeneous subsets. The splitting process results in a fully grown tree such that the value (popularity score) obtained at each terminal node (leaf node) is the mean of all label values at the node. 
\subsubsection{Gradient Boosting Decision Trees }
\label{sec:GBDT}
GBDT \cite{Ref54} is a machine learning algorithm that recursively constructs an ensemble of weak decision tree models using boosting. It has been proven to be highly efficient in various data mining competitions \cite{Ref52,Ref53}. The general principle of the GBDT algorithm is the sequential training of a series of simple decision tree estimators \cite{Ref45}, where each successive tree attempts to minimize a certain loss function formed by the preceding trees. That is, in each stage, a new regression tree is sequentially added and trained based on the residual error of the previous ensemble model. The GBDT algorithm then updates all the predicted values by adding the predicted values of the new tree. This process is recursively continued until a maximum number of trees have been generated. Thus, the final prediction value of a single instance is the sum of the predictions of all the regression trees.
 
\section{Experiments and Results}
\label{sec:Experiments}
In this section, we present the experimental setting and discuss the results.

\subsection{Experimental Setup}
\label{sec:setup}
   
\subsubsection{Popularity Measurement}
\label{sec:metric}
Social media websites allow users to interact with posted content in various ways, which results in different social signals that can be utilized to measure the popularity of social content (e.g., images, texts, and videos) on these websites. For instance, on Twitter, popularity can be gauged by the number of re-tweets, whereas the number of likes or comments can be used to measure popularity on Facebook. In this study, we use Flickr as the major image-sharing platform to predict the popularity of social media images. Previous studies have used various metrics to measure the image popularity on Flickr. For example, Khosla \textit{et al.} \cite{Ref13 } determined the popularity of an image based on the number of views it received. McParlane \textit{et al.} \cite{Ref21} adopted both view and comment count as the principal metrics.

The dataset used in our experiments complies with Khosla \textit{et al.} \cite{Ref13}, and the number of views was adopted as a popularity metric. The log function is applied to manage the large variation in the number of views for various photos from the dataset. Moreover, the images receive views during the time they are online. Thus, a log-normalization approach was used to normalize the effect of the time factor. The score proposed in \cite{Ref13} can be defined as follows: 
\begin{equation}
\label{eq:1}
\textrm{Score}_i=\textrm{log}_2 \Big(\frac{p_i}{d_i}\Big) + 1,
\end{equation}
where $p_i$ is the popularity metric (the original number of views) of image $i$, and $d_i$ is the number of days since the image first appeared on Flickr.

\subsubsection{Parameter Setting of Baseline Models}
\label{sec:Parameters}
All the baseline models were implemented using the scikit-learn machine learning library \cite{Ref55,Ref61}. In the experiments, the performance of the baseline models was observed to be significantly influenced by several hyper-parameters. Therefore,  we identified the values of a few important SVR parameters as follows: C = 3, epsilon = 0.1, gamma = auto, and kernel = RBF. Regarding the DTR model, the best performance was achieved when the max$\_$depth parameter was set to 10. Moreover, we identified several parameters of GBDT: n$\_$estimators = 2000, max$\_$depth = 10, and learning$\_$rate = 0.01. Finally, the remaining parameters are set to their default values in all the models.
\subsubsection{Dataset} 
\label{sec:A1}
\begin{figure}[htbp]
\begin{center}
\centering
\centerline{\includegraphics[width=85mm,scale=1]{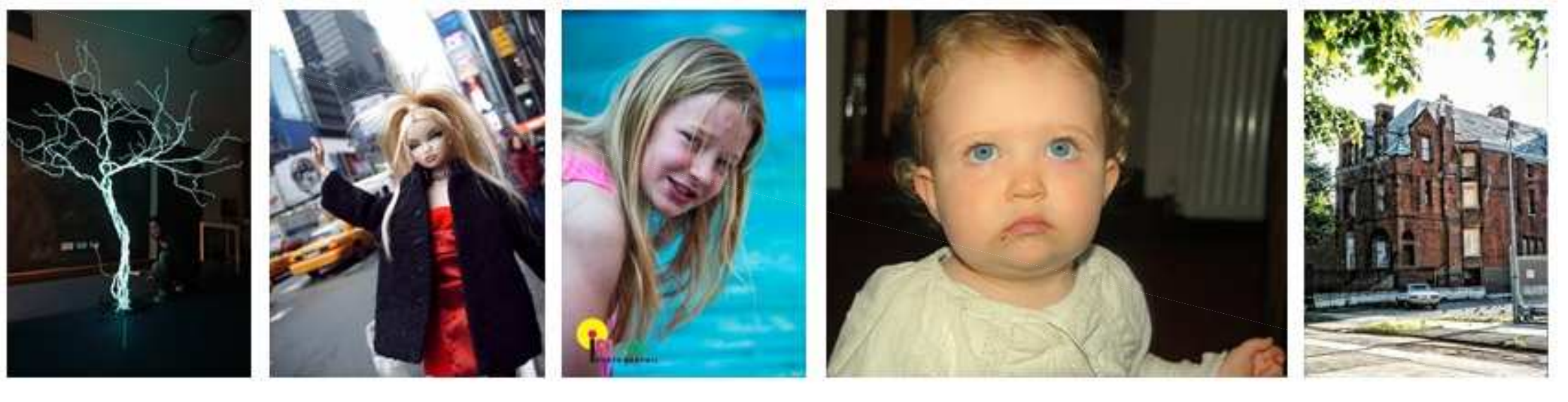}}
\caption{Sample images from the dataset. The popularity of the images is sorted from more popular (left) to less popular (right).}
\label{fig5}
\end{center}
\vspace{-2mm}
\end{figure}

The Social Media Prediction (SMP-T1) dataset presented by ACM Multimedia Grand Challenge in 2017 was used as a real-world dataset to evaluate the performance of the proposed approach \cite{Ref23,wu2017sequential}. The dataset consists of approximately 432K posts collected from the personal albums of 135 different users on Flickr. Every post in the dataset has a unique picture id along with the associated user id that signifies the user who posted the picture. Additionally, the following image metadata were provided: post date (postdate), number of comments (commentcount), number of tags in the post, whether the photo is tagged by some users or not  (haspeople), and character length of the title and image caption (titlelen or deslen). Furthermore, user-centric information, namely the average view count, group count, and average member count, was also provided in the dataset. Each image has a label representing its popularity score (log-normalized views of the image). A few images selected from the dataset are shown in Fig.~\ref{fig5}. In our experiments, 60\% of the images were used for training, 20\% for validation, and 20\% for testing.

%........Figure5.........%
\captionsetup[figure]{labelformat={default},labelsep=period,name={Fig.},skip=6pt}

%%%%%%%%%%%%%
\subsubsection{Evaluation Metrics}
\label{sec:Evaluation}
In this study, we used the same following metrics as those in the ACM Multimedia Grand Challenge \cite{Ref23,wu2017sequential} to assess the prediction accuracy: Spearman's Rho \cite{Ref43}, MSE, and mean absolute error (MAE).
\begin{itemize}
\item \textbf{Spearman's Rho:} Used to calculate the correlation between the predicted popularity scores and the actual scores for the set of tested images. \useshortskip

%\begin{equation}
%\label{eq:25}
%r_s = 1 - \frac{6\sum\limits_{j=1}^{n}d_{j}^{2}}{n(n^2-1)} 
%\end{equation}
%where $n$ is the size of the test sample and $d_j$ is the difference between the two ranks of the actual and predicted popularity values of each image $j$.

\item \textbf{MSE:} Usually used to measure the average of the sum of squared prediction errors. Each prediction error represents the difference between the actual value of the data point and the predicted value obtained by the regression model. MSE consists of simple mathematical properties, making it easier to calculate its gradient. In addition, it is often presented as a default metric for most predictive models because it is smoothly differentiable, computationally simple, and hence can be better optimized. A significant limitation of MSE is the fact that it heavily penalizes large prediction errors by squaring them. Because each error in MSE grows quadratically, the outliers in the data significantly contribute to the total error. This indicates that MSE is sensitive to outliers and applies excessive weight on their effects, which leads to an underestimation of the model performance. The drawback of MSE only becomes evident when there are outliers in the data, in which case using MAE is a sufficient alternative.

%\begin{equation}
%\label{eq:26}
%MSE = \frac{1}{n} \sum\limits_{j=1}^{n}\big(\hat{Y_j}-Y_j\big)^2
%\end{equation}
%where, in our case, $n$ is the size of the test sample, $\hat{Y_j}$ and $Y_j$ are the predicted and actual popularity scores of $j$-th image, respectively.

\item \textbf{MAE:} A simple measure usually used to evaluate the accuracy of a regression model. It measures the average of the absolute values of the individual prediction errors of the model over all samples in the test set. In the MAE metric, each prediction error contributes proportionally to the total amount of errors, indicating that larger errors contribute linearly to the overall error. Because we use the absolute value of the prediction error, the MAE does not indicate underperformance or overperformance of the model, that is, whether the regression model overpredicts or underpredicts the input samples. Thus, it offers a relatively impartial comprehension of how the model performs. By taking the absolute value of the prediction error and not squaring it, the MAE becomes more robust than MSE in managing outliers because it does not heavily penalize the large errors, as done by using MSE. Hence, MAE has its advantages and disadvantages. On one hand, it assists in handling outliers; however, on the other hand, it fails to penalize the large prediction errors.
%\useshortskip
%\begin{equation}
%\label{eq:27}
%MAE = \frac{1}{n} \sum\limits_{j=1}^{n}\abs{\hat{Y_j}-Y_j}
%\end{equation}
\end{itemize}

%\subsubsection{Implementation Platform} 
%\label{sec:A2}
%We implemented the VSCNN model using Keras \cite{Ref16}, which is a Python %library used for constructing and assessing deep learning models. Keras can %run on top of the efficient mathematical computation libraries Theano and %TensorFlow. These two libraries allow fast and easy implementation and %training of deep learning models by writing short lines of code.  

\subsection{Results} 
\label{sec:results}
Using the features extracted for model learning, we trained the proposed VSCNN model to predict the popularity score. In the training stage, we used Adam \cite{kingma2014adam} and the stochastic gradient descent as the learning optimizer to obtain the initialized parameters for VSCNN. The initial learning rate was set to 0.001. In the experiments, the model was run for 50 training epochs over the entire training set. In each epoch, the model was iterated over batches of the training set, where each batch consisted of 20 samples. Furthermore, the following features were added to the training process: 1) The learning rate was reduced by 0.1 every 10 epochs using the learning rate scheduler function, which facilitates learning. 2) The best validation accuracy was saved using the model checkpoint function, which assists in saving the best learning model. The cost function generally converges during the training phase. In the testing stage, the trained VSCNN model was applied to the test samples for evaluation. The evaluation results demonstrated that VSCNN can achieve a Spearman's Rho of 0.9014, an MAE of 0.73, and an MSE of 0.97, which are listed in Tables~\ref{tab5} and \ref{tab6}, and will be used for comparison with the baseline models.

\begin{figure}[!t]
\vspace{-2pt}
\begin{center}
\centering
\begin{subfigure}[b]{.51\linewidth}
    %\centering
    \hspace{-5mm}
    \includegraphics[width=1.05\linewidth, height=15cm, keepaspectratio,]{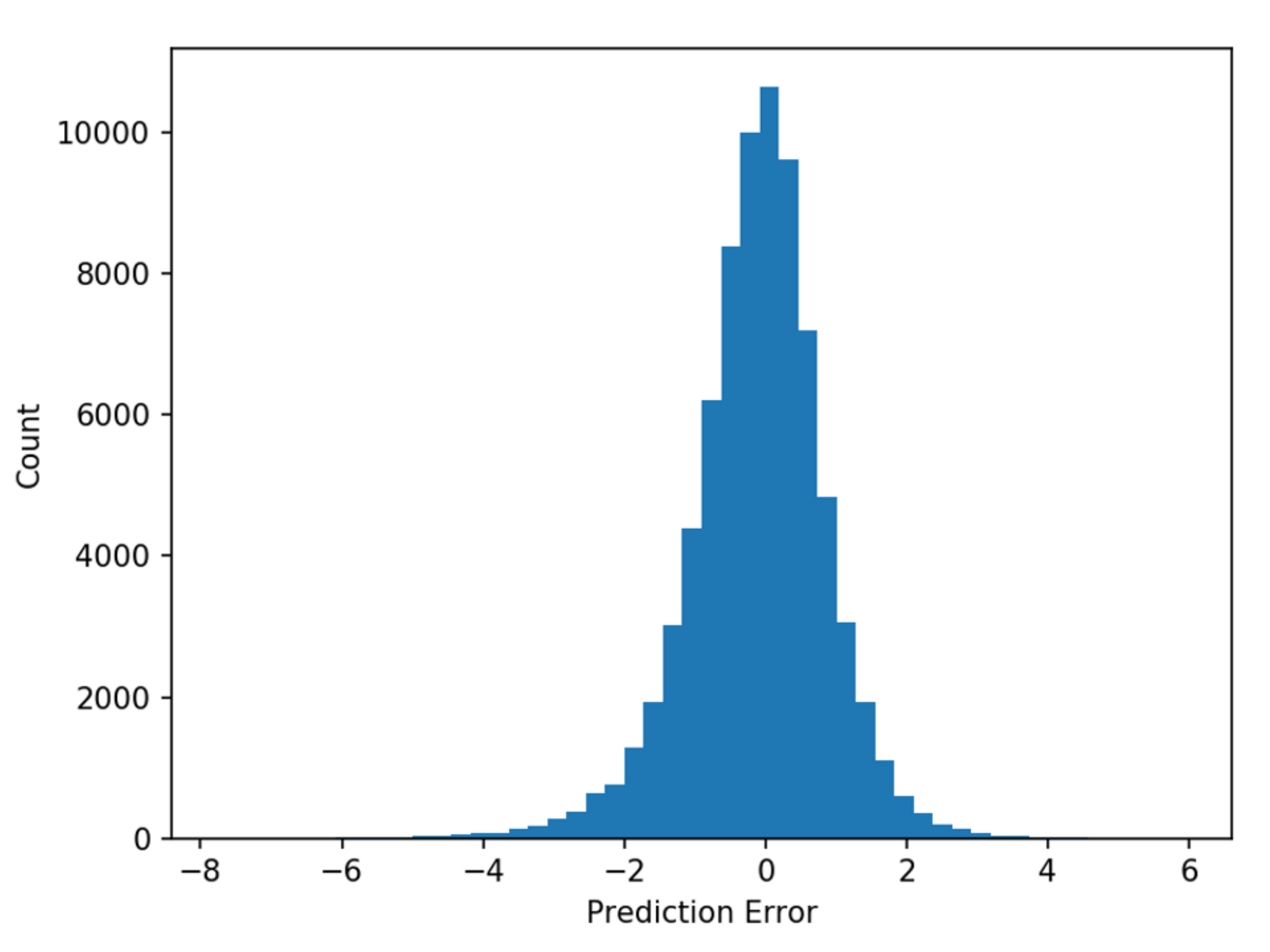} 
    %\hfill
    \captionsetup{aboveskip=1pt}
    \caption{}
    %\label{Hist}
  \end{subfigure}%   
      \begin{subfigure}[b]{.51\linewidth}
    \centering
    \hspace{-3.2mm}
    \includegraphics[width=1.05\linewidth, height=15cm, keepaspectratio,]{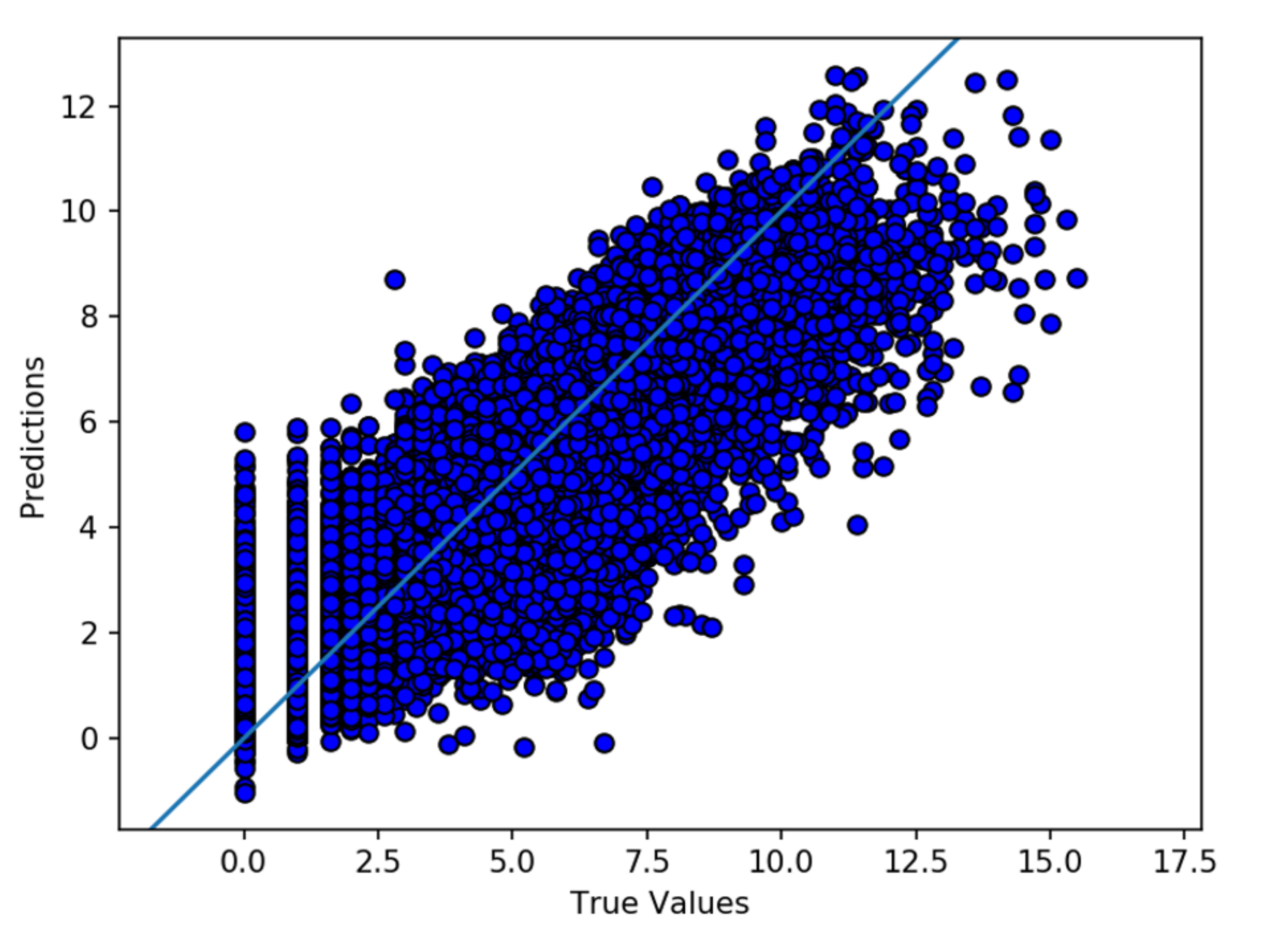}
    %\hfill
   \captionsetup{aboveskip= 1pt}
    \caption{}
    %\label{Scatter}
  \end{subfigure}% 
  \vspace{4mm}
\caption{Quality evaluation of the VSCNN model. (a) Error distribution  histogram of the model, and (b) scatterplot of true values (x-axis) versus predicted values (y-axis).}
\label{Hisandscatter}
\end{center}
\vspace{-2mm}
\end{figure}

Essential visual analytics were added for the model quality evaluation by computing the error distribution histogram, which presents the distribution of the errors made by the model when predicting the popularity score for each test sample, as shown in Fig.~\ref{Hisandscatter} (a). A larger number of errors close to zero in the histogram indicates a higher prediction accuracy. Moreover, Fig.~\ref{Hisandscatter} (b) presents a scatterplot of the actual values on the x-axis versus the predicted values obtained by the model on the y-axis. This scatterplot presents the correlation between the actual and predicted values. If the data appear to be near a straight diagonal line, it indicates a strong correlation. Thus, a perfect regression model would yield a straight diagonal line from the data. From the results shown in Fig.~\ref{Hisandscatter}, there are certain outliers that are not correctly predicted by the VSCNN model. Hence, we analyze these outliers below and explain in detail why our model fails to predict them.

\begin{figure}[htbp]
\vspace{-3mm}
\begin{center}
\centering
\centerline{\includegraphics[width=90mm,scale=1,height=5.5cm]{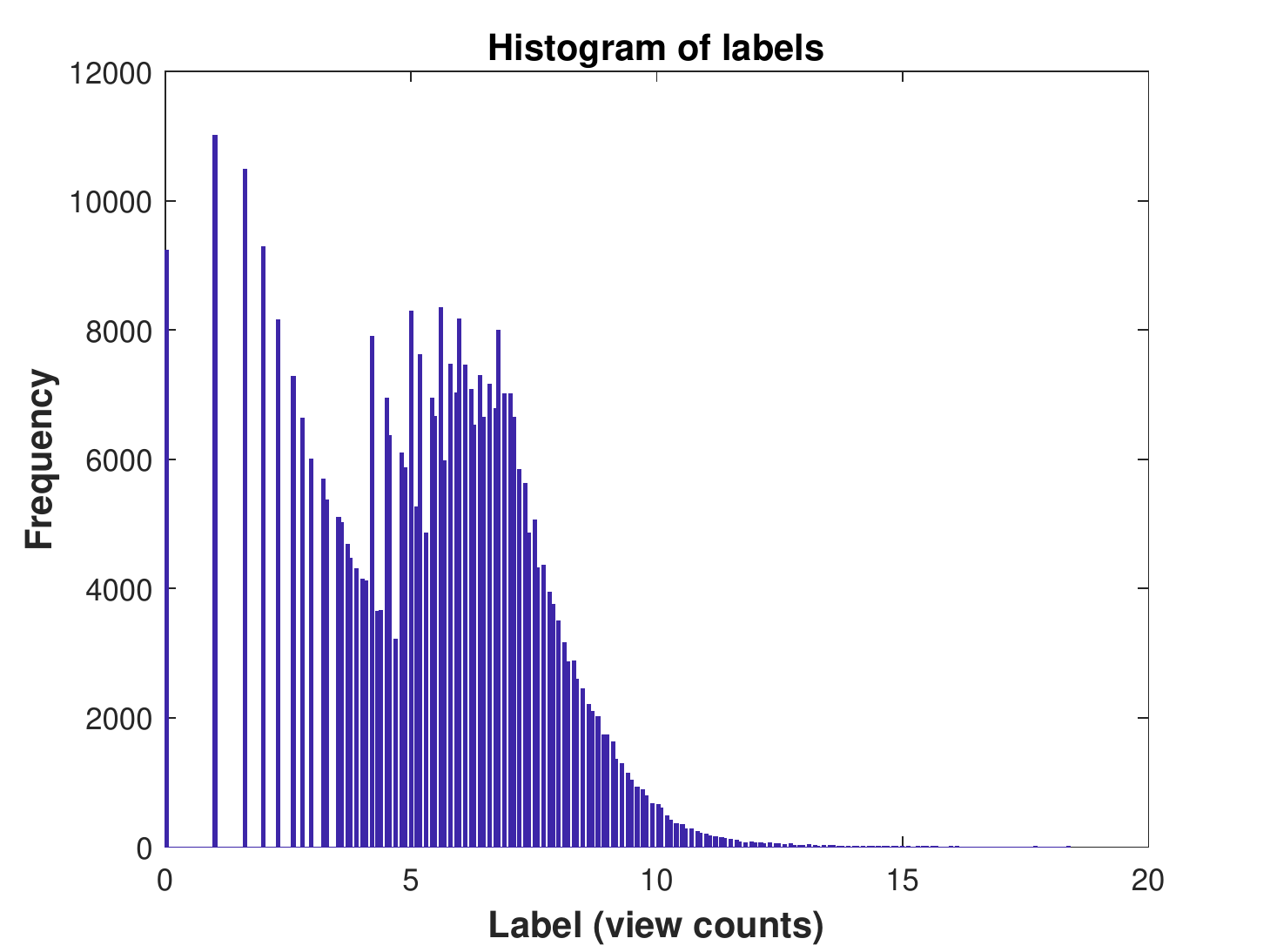}}
 \captionsetup{}

\caption{A distribution of the view counts of the training samples.}
\label{distribution}
\end{center}
\vspace{-2mm}

\end{figure}

In certain regression problems, the distribution of the target variable may have outliers (e.g., large or small values far from the mean value), which can affect the performance of the predictive model. As shown in Fig.~\ref{distribution}, the distribution of the target variable (view counts) of the training samples is highly non-uniform in our dataset; therefore, the proposed model attempts to minimize the prediction errors of the largest cluster of view counts of training samples. However, as the number of training samples with extremely high view counts is relatively low, it is more likely that the proposed model cannot correctly predict the high view counts, which will be observed as outliers in the predictive results.

\begin{figure}[ht]
\begin{subfigure}{.5\textwidth}
  \centering
  % include first image
  \includegraphics[width=.96\linewidth, height=2cm]{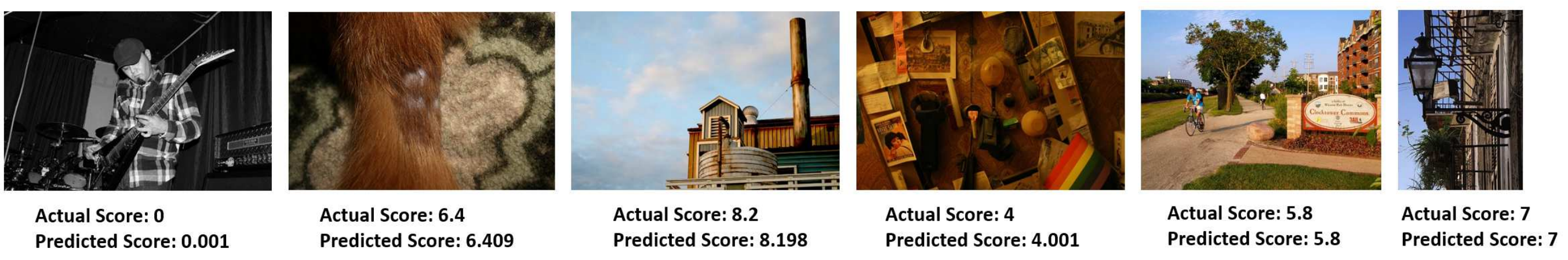} 
  \caption{Correct examples of popularity prediction}
  \label{fig:correct}
  \vspace{6mm}
\end{subfigure}
\begin{subfigure}{.49\textwidth}
  \centering
  % include second image
  \includegraphics[width=.68\linewidth, height=2cm]{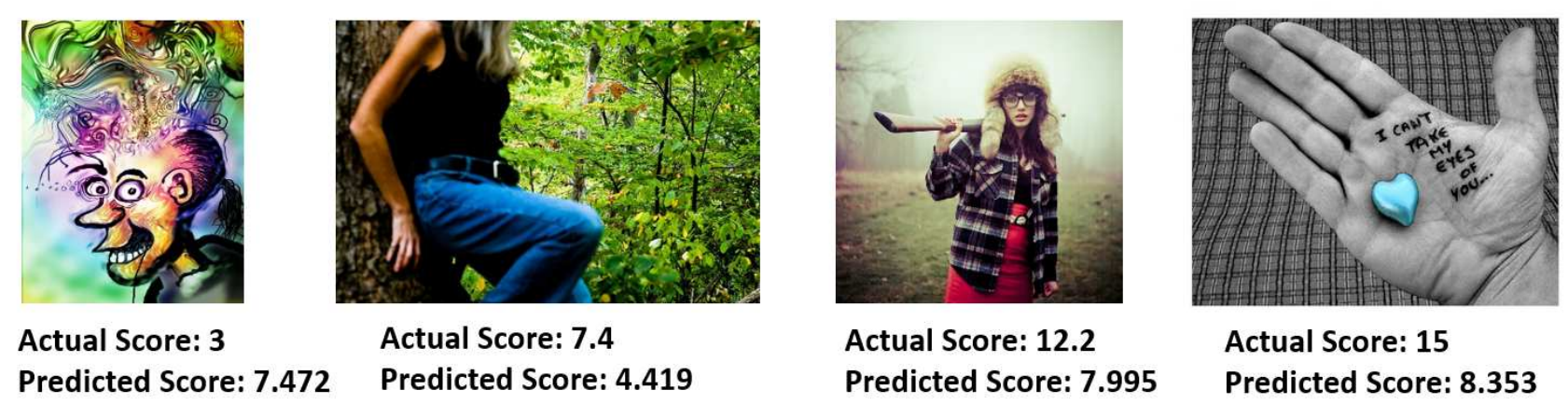}  
  \caption{Wrong examples of popularity prediction}
  \label{fig:false}
\end{subfigure}
  \vspace{5mm}
  %\captionsetup{labelfont={color=blue}} % use to color the word      Fig.8 on the graph
 \caption{Examples of correct and wrong predictions of some images from our dataset using the VSCNN model. The actual popularity score and its corresponding predicted score are displayed below each image.}

\label{fig_samples}
  \vspace{-2mm}

\end{figure}

As shown in Fig.~\ref{fig_samples}, a few good and bad predictions were made on images from the test set using our proposed model. The correctly predicted examples are shown in Fig.~\ref{fig_samples} (a); note, our model achieves superior performance with only 0.001 - 0.009 errors relative to the actual scores. For example, the popularity score of the first four images in Fig.~\ref{fig_samples} (a) is correctly predicted with errors of 0.001, 0.009, 0.002, and 0.001, respectively. In addition, the popularity score of the last two images in Fig.~\ref{fig_samples} (a) is perfectly predicted with zero prediction error. On the other hand, a few wrongly predicted examples are shown in Fig.~\ref{fig_samples} (b). For example, the actual popularity score of the first image in this figure is 3, while the score obtained by our model is 7.472, resulting in a substantial error of 4.472 in prediction. This disparity is due to the strong indications of some user features for this image, such as average views and member count, which have values of 993.42 and 10,672, respectively, and significantly contribute to the model prediction when integrating all features. Likewise, the last two images in Fig.~\ref{fig_samples} (b) are other badly predicted examples of our proposed model. The actual popularity scores of these two images are observed to be too high. Therefore, it is suggested that our model cannot correctly predict the popularity of these images because the number of training samples with high popularity scores is extremely limited in our dataset, as shown in Fig.~\ref{distribution}.

%.........Figure6........%

%%%%%%Figure7_distribution%%%%%%%%%%%
\captionsetup[figure]
{labelformat={default},labelsep=period,name={Fig.},skip=6pt}
%%%%%%%%%%%%%

%........samples Figure8.....%

\subsubsection{Comparison with Baseline Models}
\label{sec:Comparsion_with_BM}
First, we train the SCNN model using three different types of social features to explore the influence of each type on predicting popularity. Subsequently, the SCNN model is trained using all the social features as inputs. The prediction results of the SCNN model with different types of input features are summarized in Table~\ref{tab5}, which presents that the user features perform exceptionally well in predicting the popularity of an image relative to the other two types of social features (i.e., post metadata and time), with a Spearman's Rho of 0.7537, MAE of 1.13, and MSE of 2.17. This indicates that the popularity of an image is closely related to the popularity of the user uploading it; images shared on social media by popular users have a higher chance of obtaining more views. However, not all images posted by popular users are popular. To justify this, we use the popularity score and average view count as popularity metrics for images and users, respectively. As indicated in previous studies \cite{Ref21, cha2007tube}, the Pareto Principle (or 80-20 rule) was used to select a threshold to differentiate images with high (20\%) and low (80\%) popularity scores. Likewise, we set a threshold to differentiate users with a high (20\%) and low (80\%) average view count. Based on these differentiations, the top 20\% of the images and users are considered as highly popular (or popular), while the remaining 80\% are considered less popular (or common). Accordingly, on average, 69.19\% and 16.17\% of the images posted by popular and common users, respectively, were determined to be popular. Thus, we conclude that not all images posted by popular users are always popular.

The post metadata are also noteworthy features. The SCNN model using these features achieved values of 0.6590, 1.35, and 2.98 for the Spearman’s Rho, MAE, and MSE, respectively. This indicates that image-specific social features, such as tag count, title length, description length, and comment count, also play an important role in predicting popularity, which is expected; an image with significant tags or a longer description/title tends to be more popular because it has a greater chance of showing up in the search results when people use keywords to search for images. Similarly, having more comments on the image suggests that more users interact with the image, which may lead to a greater number of views and thus, increased popularity. Considering the results, time features were also determined to make a significant contribution to popularity prediction, which indicates that the time when an image is posted may influence its popularity. For example, users tend to browse social networking sites at a particular time of the day, such as weekend leisure time, which indicates that images posted during that time are more likely to receive a large number of views and become popular.

\begin{figure}[!t]
\vspace{1mm}
\begin{center}
\centering
\begin{subfigure}[t]{.51\linewidth}
    %\centering
    \hspace{-4mm}
    \includegraphics[width=1.01\linewidth, height=15cm, keepaspectratio,]{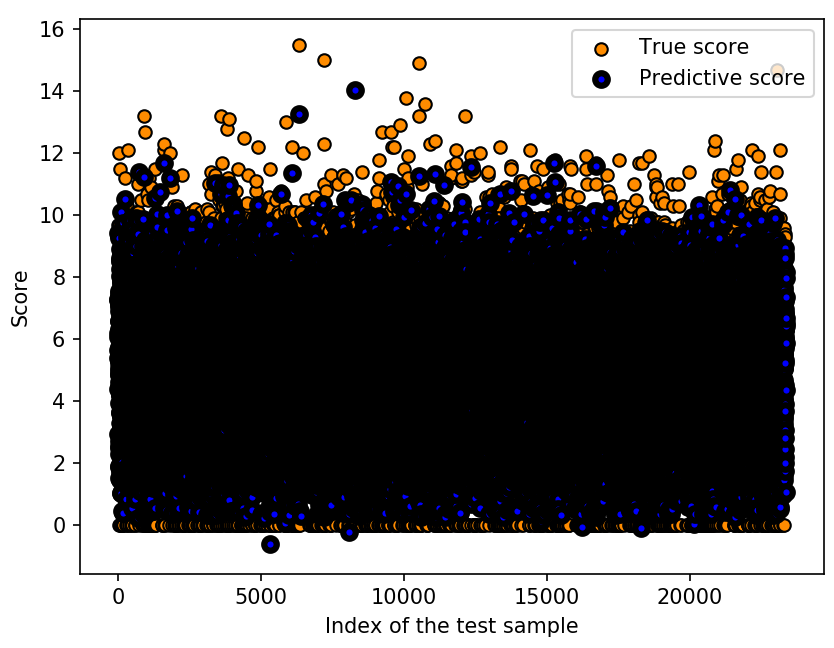}
    \captionsetup{aboveskip=2pt}
    \caption{}
    %\label{Hist}
  \end{subfigure}%   
  \begin{subfigure}[t]{.51\linewidth}
    %\centering
    \hspace{-3.1mm}
    \includegraphics[width=1\linewidth, height=15cm, keepaspectratio,]{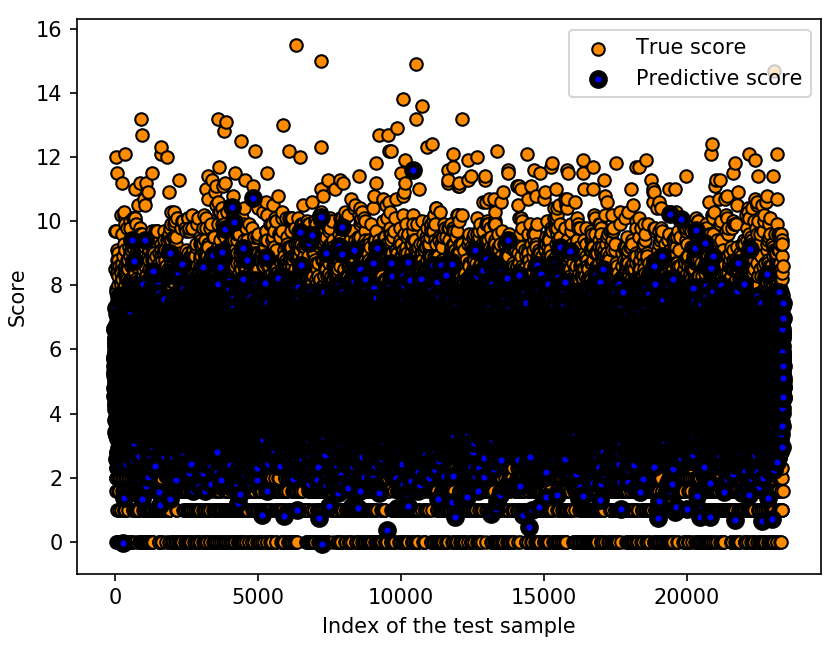}
    \captionsetup{aboveskip=2pt}
    \caption{}
    %\label{Scatter}
  \end{subfigure}%
     \vspace{2mm}
%\captionsetup{labelfont={color=blue}}
\caption{Diagrams of the predicted values obtained using the CNN-based baseline models and their corresponding ground truth values. (a) SCNN. (b) VCNN.}
\label{CNN_digrams}
\end{center}
\vspace{-2mm}
\end{figure}

Furthermore, while each type of social feature performs sufficiently, the SCNN model achieves the best predictive performance when all the social features are combined, as shown in the fourth row of Table~\ref{tab5}. This suggests that all the social features proposed are strongly correlated and provide complementary information to each other. Fig.~\ref{CNN_digrams} (a) presents a diagram of the predicted values obtained by SCNN and the corresponding actual values.

Similarly, the VCNN model was trained using each of the individual visual features to analyze their effect on predicting image popularity. We also integrated all the visual features and used them as the input to the model. The evaluation results are listed in Table~\ref{tab5}. Deep learning features were observed to outperform other visual features. However, it is important to note that the VCNN model achieves the best performance in terms of all the evaluation metrics when all the visual features are combined. In addition, as indicated by the results of the VCNN model, visual features are less effective than social features in terms of image popularity prediction. This finding is consistent with previous studies \cite{Ref13, lv2017multi, wang2017combining, huang2017towards}. Nevertheless, the visual features are useful when there are no post metadata existing, or to address scenarios where no social interactions were recorded prior to publishing the image (e.g., user newly joined social network). This indicates that image content also plays a critical role in popularity prediction, and may complement the social features.

A diagram of the predicted values obtained using VCNN and the corresponding actual values is shown in Fig.~\ref{CNN_digrams} (b). Finally, the performance of the proposed model is compared with the best performance of both VCNN and SCNN in terms of all the evaluation metrics; the results are listed in Table~\ref{tab5}. Apparently, VSCNN outperforms VCNN and SCNN, with a relative improvement of 2.33\% (SCNN) and 116.27\% (VCNN) in terms of Spearman's Rho, and a decrease of 7.59\% (SCNN) and 53.80\% (VCNN), as well as 14.16\% (SCNN) and 76.23\% (VCNN) in terms of MAE and MSE, respectively. 

%%% Table4%%%%%
\begin{table}
\caption{Performance comparison of SCNN, VCNN, and VSCNN  models.  }
\label{tab5}%\setlength{\tabcolsep}{3pt}
\centering
\begin{tabular}{|P{30pt}|P{50pt}|P{55pt}|P{20pt}|P{20pt}|}
%\begin{tabular}{|c|c|c|c|c|}
\hline
Models& Features&
Spearman's Rho& 
MAE &
MSE
 \\
\hline
SCNN& User&0.7537&1.13&2.17\\ 
&Post\_Metadata&0.6590&1.35&2.98\\
&Time&0.5317&1.42&3.43\\
 %\rowcolor{blue}
&All\_Social&0.8809&0.79&1.13\\
\hline
VCNN&Color&0.3278&1.66&4.46\\
&Gist&0.2612&1.72&4.67\\
&LBP&0.3287&1.66&4.45\\
&Aesthetic&0.2000&1.77&4.91\\
&Deep&0.4101&1.61&4.13\\
&All\_Visual&0.4168&1.58&4.08\\
\hline
VSCNN&Visual+Social&\textbf{0.9014}&\textbf{0.73}&\textbf{0.97}\\
\hline
\end{tabular}
\newline
\vspace{-2mm}
\end{table}

%........Figure9........%

%.......Figure10.......%
\begin{figure}[!t]
\begin{center}
\vspace{-2mm}
\centering
\begin{subfigure}[t]{.51\linewidth}
    %\centering
    \hspace{-4mm}
    \includegraphics[width=1.01\linewidth, height=15cm, keepaspectratio,]{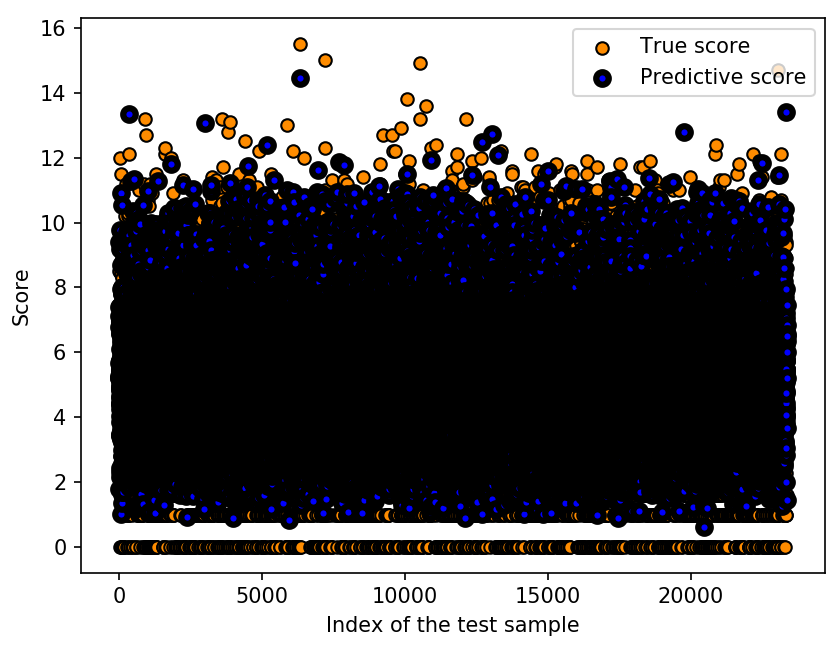}
    \captionsetup{aboveskip=2pt}
    \caption{}
    %\label{Hist}
    \vspace{5mm}

  \end{subfigure}%   
  \begin{subfigure}[t]{.51\linewidth}
    %\centering
    \hspace{-3.1mm}
    \includegraphics[width=1\linewidth, height=15cm, keepaspectratio,]{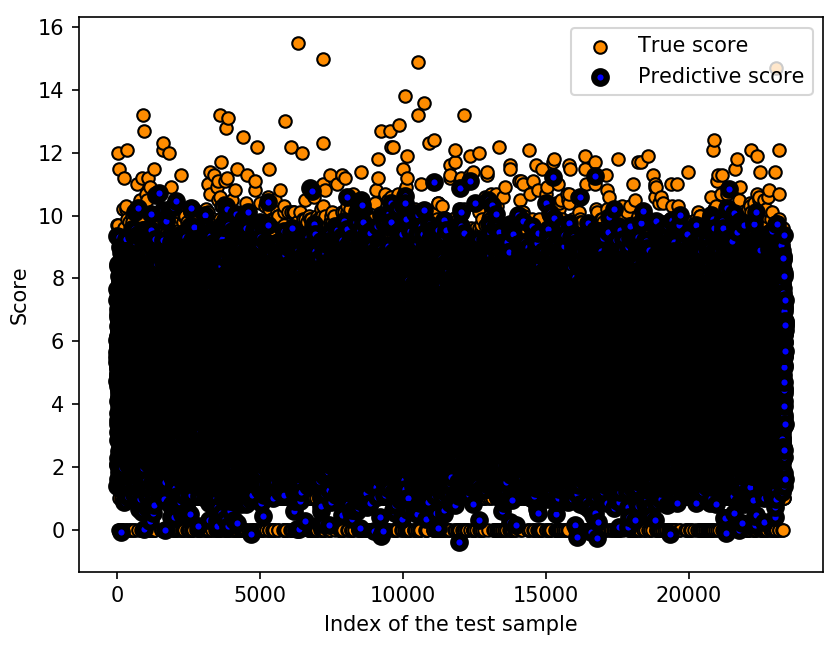}
    \captionsetup{aboveskip=2pt}
    \caption{}
    %\label{Scatter}
  \end{subfigure}%
 
  \begin{subfigure}[t]{.51\linewidth}
    %\centering
    \hspace{-4mm}
    \includegraphics[width=1.01\linewidth, height=15cm, keepaspectratio,]{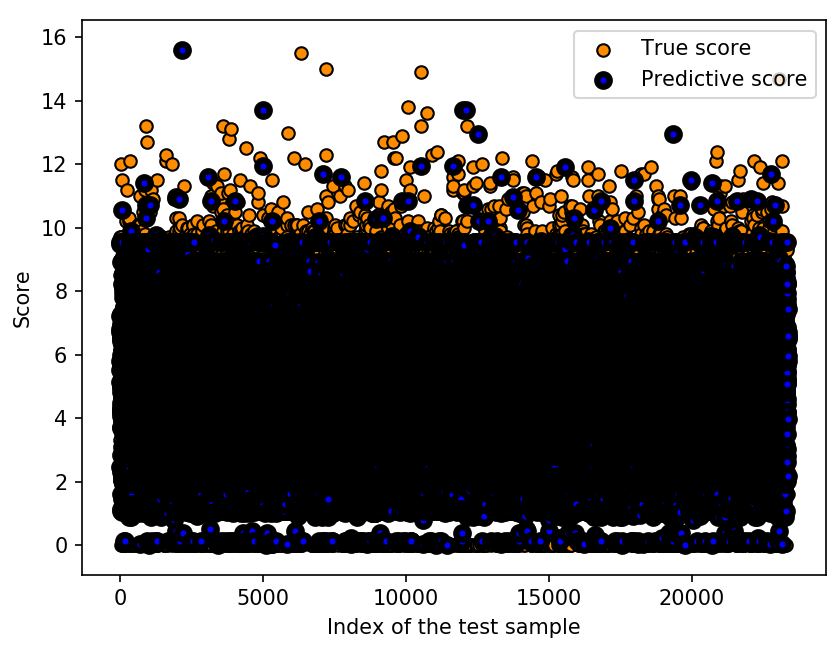}
    \captionsetup{aboveskip=2pt}
    \caption{}
    %\label{Scatter}
  \end{subfigure}% 
  \begin{subfigure}[t]{.51\linewidth}
    %\centering
    \hspace{-3.1mm}
    \includegraphics[width=1\linewidth, height=15cm, keepaspectratio,]{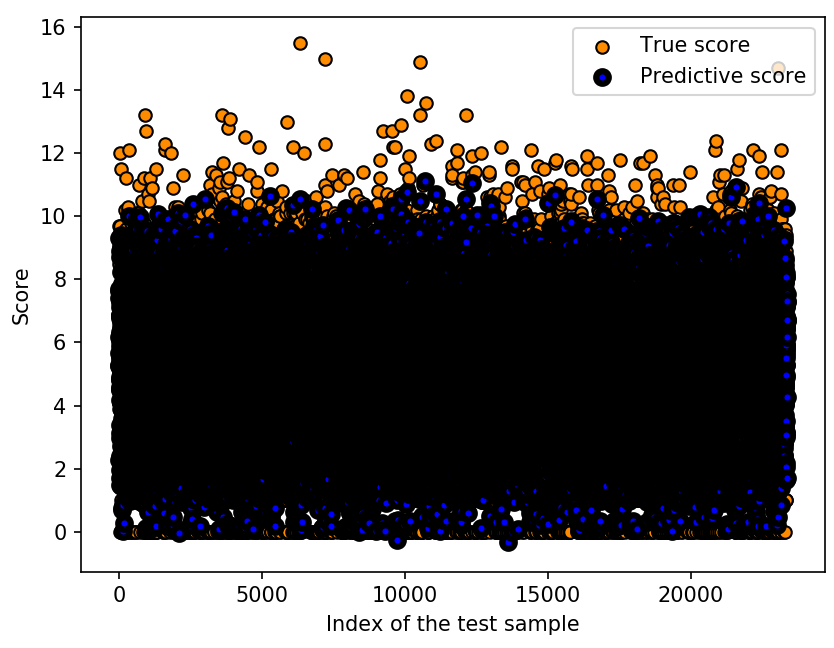}
    \captionsetup{aboveskip=2pt}
    \caption{}
        
    %\label{Scatter}
  \end{subfigure}% 
      \vspace{2mm}
\captionsetup{}
 \caption{Diagrams of the predicted values obtained using the four machine learning baseline models and their corresponding ground truth values. (a) LR, (b) SVR, (c) DTR, and (d) GBDT.}
\label{digrams}
\end{center}
\vspace{-2mm}
\end{figure}

Subsequently, the other four baseline models (i.e., LR, SVR, DTR, and GBDT) were trained using each single feature and various combinations thereof to demonstrate the effectiveness of the proposed features in predicting image popularity. The predictions are shown in Table~\ref{tab6}, presenting that the user feature yields the best results. This indicates that the characteristics of the person who posts a photo determine its popularity to a significant extent. Furthermore, post metadata and time features were also determined to be sufficient predictors. Additionally, when all social context features are combined and used as inputs, the performance improves significantly for all the models, and GBDT achieves the best performance in terms of all the evaluation metrics.

The deep learning feature is significant and outperforms other visual features; namely, color, gist, LBP, and aesthetics, although these features perform sufficiently in all models. Nevertheless, the performance of all models is improved when all visual features are combined. Moreover, note that combining visual and social features leads to a significant improvement in the performance of all the models compared to that exhibited using either set of these features independently.

Fig.~\ref{digrams} presents diagrams of the predicted values obtained using the four machine learning baseline models and their corresponding ground truth values. As presented in Table~\ref{tab6} and shown in Fig.~\ref{digrams}, GBDT outperformed all other machine learning models, with a relative improvement from 5.16\% (SVR) to 19.57\% (LR) in terms of Spearman's Rho, and with decreases from 13.98\% to 34.43\% and from 25.32\% to 52.87\% in terms of MAE and MSE, respectively. Finally, the performance of the proposed VSCNN model was compared with the best performance obtained by each of the four baseline models (LR, SVR, DTR, and GBDT); the results are shown in Table~\ref{tab6}. Compared with GBDT, VSCNN improves the prediction performance by approximately 2.69\%, 8.75\%, and 15.65\% in terms of Spearman's Rho, MAE, and MSE, respectively.

%%% Table5%%%%%
\begin{table*}
\caption{Performance comparison of LR, SVR, DTR, GBDT, and VSCNN models.  }
\label{tab6}%\setlength{\tabcolsep}{3pt}
\centering
\begin{tabular}{|P{43pt}|P{32pt}|P{13pt}|P{12pt}|P{32pt}|P{13pt}|P{12pt}|P{32pt}|P{13pt}|P{12pt}|P{32pt}|P{13pt}|P{12pt}|P{32pt}|P{13pt}|P{12pt}|}
%\begin{tabular}{|c|c|c|c|c|c|c|c|c|c|c|c|c|c|c|c}
\hline
\multirow{2}{*}{Features}& \multicolumn{3}{c|}{LR}&\multicolumn{3}{c|}{SVR}&\multicolumn{3}{c|}{DTR}&\multicolumn{3}{c|}{GBDT}&\multicolumn{3}{c|}{VSCNN}\\
 \cline{2-16}
& Spearman's Rho & MAE & MSE & Spearman's Rho                                & MAE & MSE & Spearman's Rho & MAE & MSE& Spearman's Rho & MAE & MSE& Spearman's Rho & MAE & MSE\\
\hline
Color&0.0856&1.81&5.10&0.2569&1.72&4.74&0.1915&1.76&4.93&0.3381&1.66&4.41&&&\\
Gist&0.1337&1.79&5.04&0.3209&1.67& 4.55&0.1436&1.80& 5.12&0.3176&1.69&4.51&&&\\
LBP&0.1546&1.79&5.04&0.3028&1.69&4.63&0.1640&1.78&5.01&0.3126&1.68&4.52&&&\\
Aesthetic&0.1221&1.8&5.06&0.1866&1.77&4.97&0.1661&1.79&5.00&0.2040&1.77&4.88&&&\\
Deep&0.3701&1.66&4.43&0.4754&1.53&3.88&0.2330&1.76&4.96&0.4403&1.59&4.05&&&\\
All\_Visual&0.3837&1.65&4.35&0.5018&1.50&3.73&0.2384&1.76&4.95&0.4890&1.53&3.82&&&\\
user&0.6449&1.41&3.17&0.7548&1.12&2.18&0.7579&1.12&2.15&0.7580&1.12&2.15&&&\\
Post\_Metadata&0.5266&1.68&4.41&0.6126&1.42&3.28&0.6682&1.32&2.91&0.6962&1.27&2.72&&&\\
Time&0.1337&1.80&5.00&0.2681&1.70&4.65&0.3485&1.61&4.19&0.6285&1.29&2.84&&&\\
All\_Social&0.7114&1.28&2.72&0.8292&0.94&1.58&0.8049&1.01&1.74&0.8611&0.86&1.30&&&\\
Visual+Social&0.7341&1.22&2.44&0.8347&0.93&1.54&0.8200&0.96&1.65&0.8778&0.80&1.15&\textbf{0.9014}&\textbf{0.73}&\textbf{0.97}\\
\hline
\end{tabular}
\newline
\vspace{3mm}
\end{table*}

\begin{figure}[htbp]
\vspace{-5pt}
\begin{center}
\centering
\centerline{\includegraphics[width=90mm,scale=1, height=6cm]{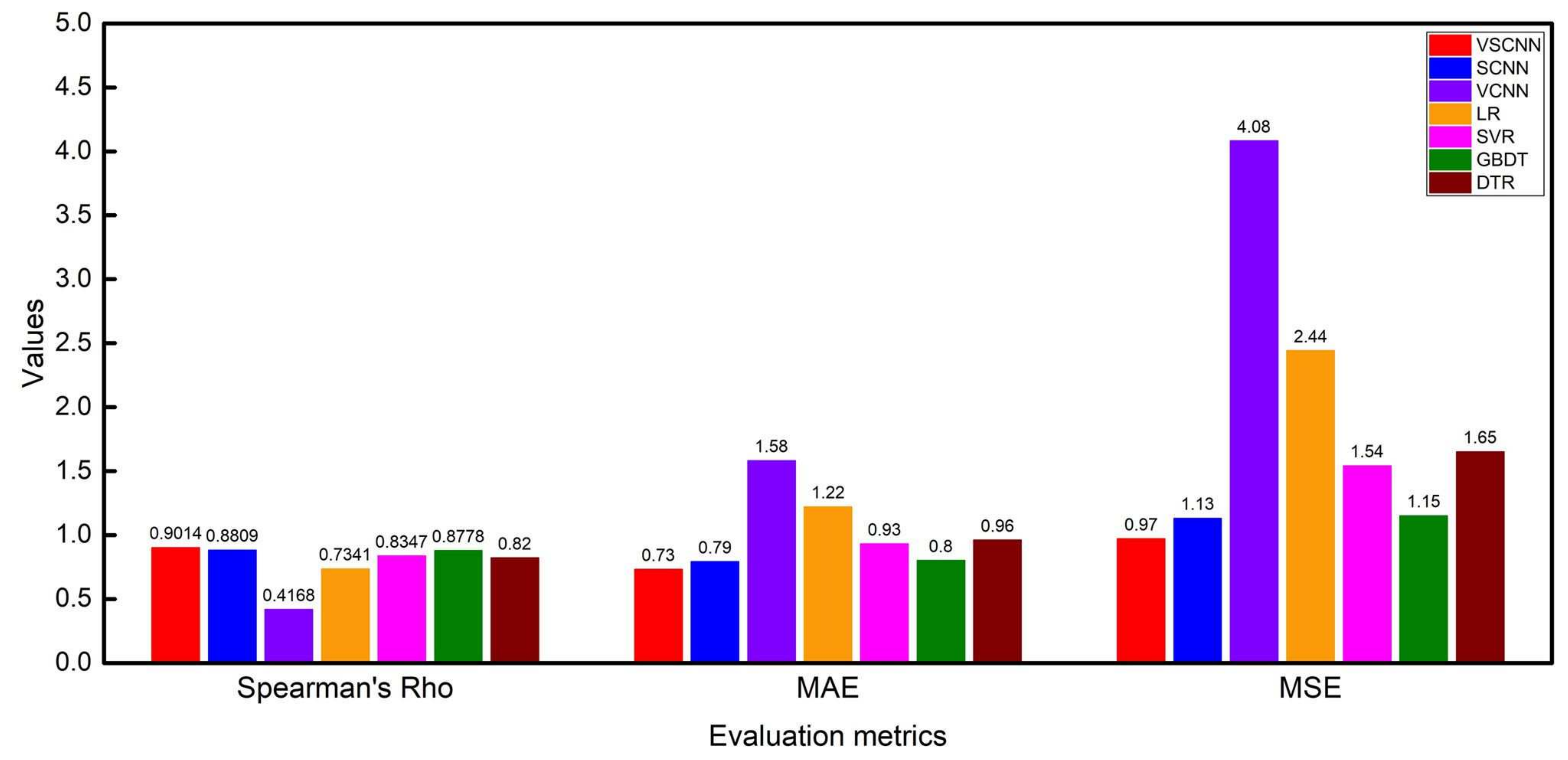}}
\caption{Best prediction performances for all the models in terms of Spearman's Rho, MAE, and MSE metrics.}
\label{Bestperformance}
\end{center}
\vspace{-2mm}
\end{figure}

Fig.~\ref{Bestperformance} presents the best prediction performance for all the models in terms of the three evaluation metrics; the VSCNN outperforms all six baseline models in predicting the popularity of an image. Overall, VSCNN achieved the best prediction performance, with the highest Spearman's Rho (0.9014) and lowest MAE and MSE (0.73 and 0.97, respectively). This suggests that CNNs are more powerful than other machine learning methods in processing heterogeneous information for popularity prediction. Another significant finding is that both social and image content features are essential and complement each other in predicting image popularity on photo-sharing websites.

%.......Figure11......%
%%%%%%%%%%%%%

%%% Table6%%%%%
\begin{table}
\caption{Comparison with the state-of-the-art methods on SMP-T1 dataset.  }
\label{SMP-T1-models}%\setlength{\tabcolsep}{3pt}
\centering
\begin{tabular}{|P{13pt}|P{107pt}|P{32pt}|P{17pt}|P{21pt}|}
%\begin{tabular}{c|c|l|l|l}
\hline
\multicolumn{2}{|c|}{Methods}& Spearman's Rho& MAE& MSE
 \\
\hline

\parbox[t]{2mm}{\multirow{9}{*}{\rotatebox[origin=c]{90}{\parbox{2.5cm}{SMP Challenge Teams \centering \cite{Ref23,wu2017sequential}}}}}&TaiwanNo.1 SMP-T1&0.8268&1.0676&2.0528\\
&heihei SMP-T1&0.8093&1.1059&2.1767\\
&NLPR\_MMC\_Passerby SMP-T1&0.7927&1.1783&2.4973\\
&BUPTMM SMP-T1&0.7723&1.1733&2.4482\\
&bluesky SMP-T1&0.7406&1.2475&2.7293\\
&WePREdictIt SMP-T1&0.5631&1.6278&4.2022\\
&FirstBlood SMP-T1&0.6456&1.6761&6.3815\\
&ride\_snail\_to\_race SMP-T1&-0.0405&2.4274&9.2715\\
&CERTH-ITI-MKLAB SMP-T1&0.3554&3.8178&19.3593\\
\hline
\multicolumn{2}{|c|}{Multimodal approach \cite{meghawat2018multimodal}}&0.75&1.12&2.39\\
\hline
\multicolumn{2}{|c|}{VSCNN}&\textbf{0.9014}&\textbf{0.73}&\textbf{0.97}\\
\hline
\end{tabular}
%\newline
\vspace{4mm}
\end{table}

%%% Table7%%%%%
\begin{table}
\caption{Performance comparison of VSCNN and VSCNN-EF models.  }
\label{tab4}%\setlength{\tabcolsep}{3pt}
\centering
\begin{tabular}{|P{50pt}|P{60pt}|P{30pt}|P{30pt}|}
%\begin{tabular}{c|c|c|c}
\hline
Models& 
Spearman's Rho& 
MAE &
MSE
 \\
\hline
VSCNN&\textbf{0.9014}&\textbf{0.73}&\textbf{0.97}\\
\hline
VSCNN-EF&0.8898&0.76&1.07\\

\hline
\end{tabular}
\newline
\vspace{-4.5mm}
\end{table}

\subsubsection{Comparison with state-of-the-art methods on SMP-T1 dataset}
\label{sec:Comparsion_with_challenge}
The ACM Multimedia Grand Challenge in 2017 presented a social media prediction task (SMP-T1) as a challenge \cite{Ref23,wu2017sequential}, to predict the popularity of images posted by users on social media. Several teams participated in this challenge, and proposed different models based on the provided SMP-T1 dataset. We compared the performance of the proposed VSCNN model with that of these models, for which the evaluation results are listed in Table~\ref{SMP-T1-models}; VSCNN outperforms all the other models. Compared to the best team model (i.e., TaiwanNo.1 SMP-T1), VSCNN improves the prediction performance by approximately 9.02\%, 31.62\%, and 52.75\% in terms of Spearman's Rho, MAE, and MSE, respectively.

In addition to the aforementioned compared models, the multimodal approach presented in \cite{meghawat2018multimodal} integrates significant multimodal information extracted from the same SMPT1 dataset into a CNN model for predicting the popularity of images. Although this approach adopts multimodal features (e.g., image, textual, contextual, and social features) for popularity prediction, it ignores other important features, such as low-level computer vision, aesthetics, and time features. It also adopts an early fusion scheme to merge the features extracted from various modalities into a single large feature vector prior to feeding them into the CNN regression model. Although early fusion can create a joint representation of the input features from multiple modalities, it requires the features to be extremely engineered and preprocessed to be aligned well before the fusion process. It also suffers from the difficulty of representing the time synchronization between multimodal features. Moreover, the increase in the number of modalities makes it difficult to learn the cross-correlation among the overly heterogeneous features. Eventually, a single model is used to make predictions by assuming that the model is well suited for all modalities. However, the architecture of the CNN model used in \cite{meghawat2018multimodal} is not sufficiently powerful to process features from different modalities and then accurately predict the popularity of an image.

Unlike [32], our model employs a late fusion scheme in which the features of each modality (i.e., visual and social features) are examined and trained independently using two CNNs with a highly designed architecture. The obtained results are then fused using a merged layer into another network for further processing and obtaining the final prediction. The fusion process in our model becomes easy to execute and does not suffer from the data representation problem that the early fusion scheme has because the semantic vectors resulting from the two CNN models usually have the same form of data. In addition, late fusion allows the usage of the most suitable model for analyzing each modality and learning its features, providing more flexibility. Furthermore, the robust interpretation of incomplete and inconsistent multimodal input becomes more reliable at later stages because more semantic knowledge becomes available from various sources. Owing to these advantages, the late fusion scheme is extensively used in multimodal systems \cite{ Ref13,hou2018audio,lv2017multi}. To confirm the efficiency of our model, we also compared it to the multimodal approach proposed in \cite{meghawat2018multimodal}; the prediction results are summarized in Table~\ref{SMP-T1-models}. Apparently, VSCNN outperforms the multimodal approach, with a relative improvement of 20.19\% in terms of Spearman's Rho, and a decrease of 34.82\% and 59.41\% in terms of MAE, and MSE, respectively.

\subsubsection{Late and Early Fusion Schemes for the VSCNN model}
\label{sec:EarlyandLate}
The framework proposed in this study adopts the late fusion scheme; that is, we first employ two convolutional neural networks to process visual features and the corresponding social context information individually. Then, the outputs of these two networks are merged into another network that fully connects all the information into a final layer of the deep architecture. We also tested the early fusion scheme by integrating visual and social features at the input of the convolutional layers. The early fusion scheme, denoted by VSCNN-EF, replaces the visual and social networks in Fig.~\ref{framework} with a unified CNN whose inputs comprise of fused visual-social features obtained by concatenating the visual and social features of a given image into a final feature vector of 4,744 dimensions (4,710 visual and 34 social). Then, PCA \cite{jolliffe2016principal} is applied to reduce the dimensionality of this vector from 4,744 to 20 and to select only the most prevalent features. The numbers of parameters of VSCNN and VSCNN-EF are of the same order. These schemes are optimized and tested, followed by comparing the prediction performance in terms of the three performance metrics; the results are listed in Table~\ref{tab4}. Apparently, VSCNN consistently outperforms VSCNN-EF, suggesting that the proposed late fusion scheme, which initially processes visual and social information independently and merges them later, is better than an early fusion scheme, which incorporates the heterogeneous data at the beginning.

\section{Conclusion}
\label{sec:conc}
Recently, deriving an effective computational model to characterize human behavior or predict decision making has become an emergent topic. In this study, we developed a multimodal deep learning framework for predicting the popularity of images on social media. First, we analyzed and extracted different types of image visual content features and social context information that significantly affect image popularity. Then, we proposed a novel CNN-based visual-social computational model for image popularity prediction, called VSCNN. This model uses individual networks to process input data with different modalities (i.e., visual and social features), and the outputs from these networks are then integrated into a fusion network to learn joint multimodal features and estimate the popularity score. We trained the proposed model in an end-to-end manner. The experimental results on the provided dataset demonstrate the effectiveness of the proposed model in predicting image popularity. Further experiments demonstrated that VSCNN achieved a notably superior prediction performance. Specifically, it outperformed four traditional machine learning schemes, two CNN-based models, and other state-of-the-art methods, in terms of three standard evaluation metrics (i.e., Spearman's Rho, MAE, and MSE). This emphasizes the effectiveness of the proposed model in combining visual and social information to predict the popularity of an image. 

In the future, we will extend our work by considering not only internal, but also external factors that may affect image popularity, such as real-world events. Meanwhile, we will investigate the influence of various aspects on image popularity based on geographical location and cultural background. Additionally, we plan to use a generative model as suggested in \cite{vinyals2016show} to automatically generate natural sentences describing the content and title for each image in the SMP-T1 dataset, and using an image annotation model as proposed in \cite{wu2018tagging} to create a set of keywords (hashtags) that are related to the content of the image. We then incorporate the obtained textual information into our model to explore its effect on image popularity. Finally, we aim to optimize the parameters and overall structures of the CNNs used in the proposed model to improve prediction performance.

\section*{Acknowledgement}

This work was supported in part by the Ministry of Science and Technology of Taiwan under Grant MOST 109-2634-F-008-006 -. MOST 109-2221-E-001-016 -.

% or


%\appendices

%\section{}

%\section*{Acknowledgment}


\ifCLASSOPTIONcaptionsoff
  \newpage
\fi



%\bibliographystyle{IEEEtran}

%\bibliographystyle{IEEEtran}
%
%\bibliography{IEEEabrv,reference}
%\bibliographystyle{IEEEtran}
%\bibliography{IEEEabrv.bib,reference.bib}{}
%\bibliography{reference.bib}{}
%\bibliographystyle{plain}

\bibliographystyle{IEEEtran}
\bibliography{reference}

%\vspace*{-4\baselineskip}
%
%\vspace{1cm}
%
%
%\vskip -2\baselineskip plus -1fil

\begin{IEEEbiography}[{\includegraphics[width=1in,height=1.25in,clip,keepaspectratio]{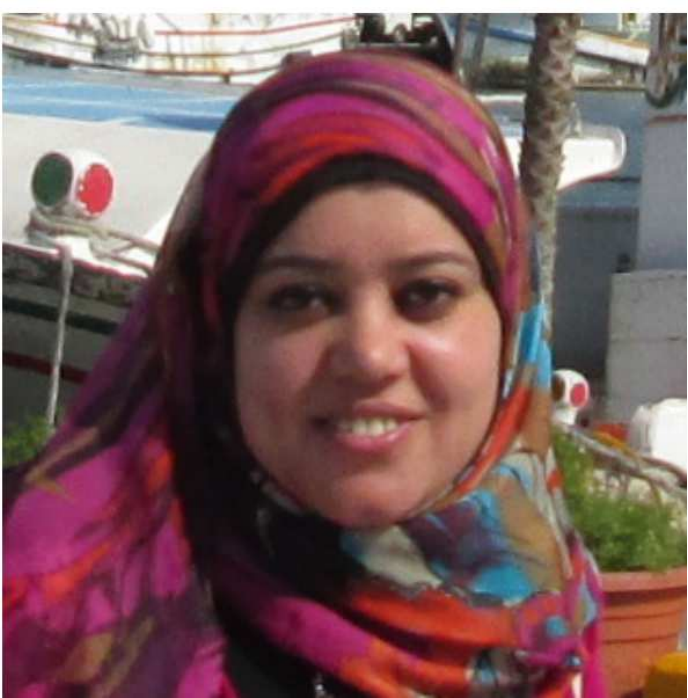}}]{Fatma S. Abousaleh received the B.S. degree in mathematics and computer science from Zagazig University, Zagazig, Egypt, in 2003, and the M.S. degree in computer science from Ain Shams University, Cairo, Egypt, in 2010. She is currently pursuing the Ph.D. degree with the Taiwan International Graduate Program in Social Networks and Human-Centered Computing, Institute of Information Science, Academia Sinica, Taipei, Taiwan, and the Department of Computer Science, Faculty of Science, National Chengchi University, Taipei, Taiwan. Her research interests include image processing and social network analysis.}
\end{IEEEbiography}
%\vskip -2pt plus -2fil
%\vspace*{-3.5\baselineskip}
%\vspace{-3cm}
\vskip -2\baselineskip plus -1fil

\begin{IEEEbiography}[{\includegraphics[width=1in,height=1.25in,clip,keepaspectratio]{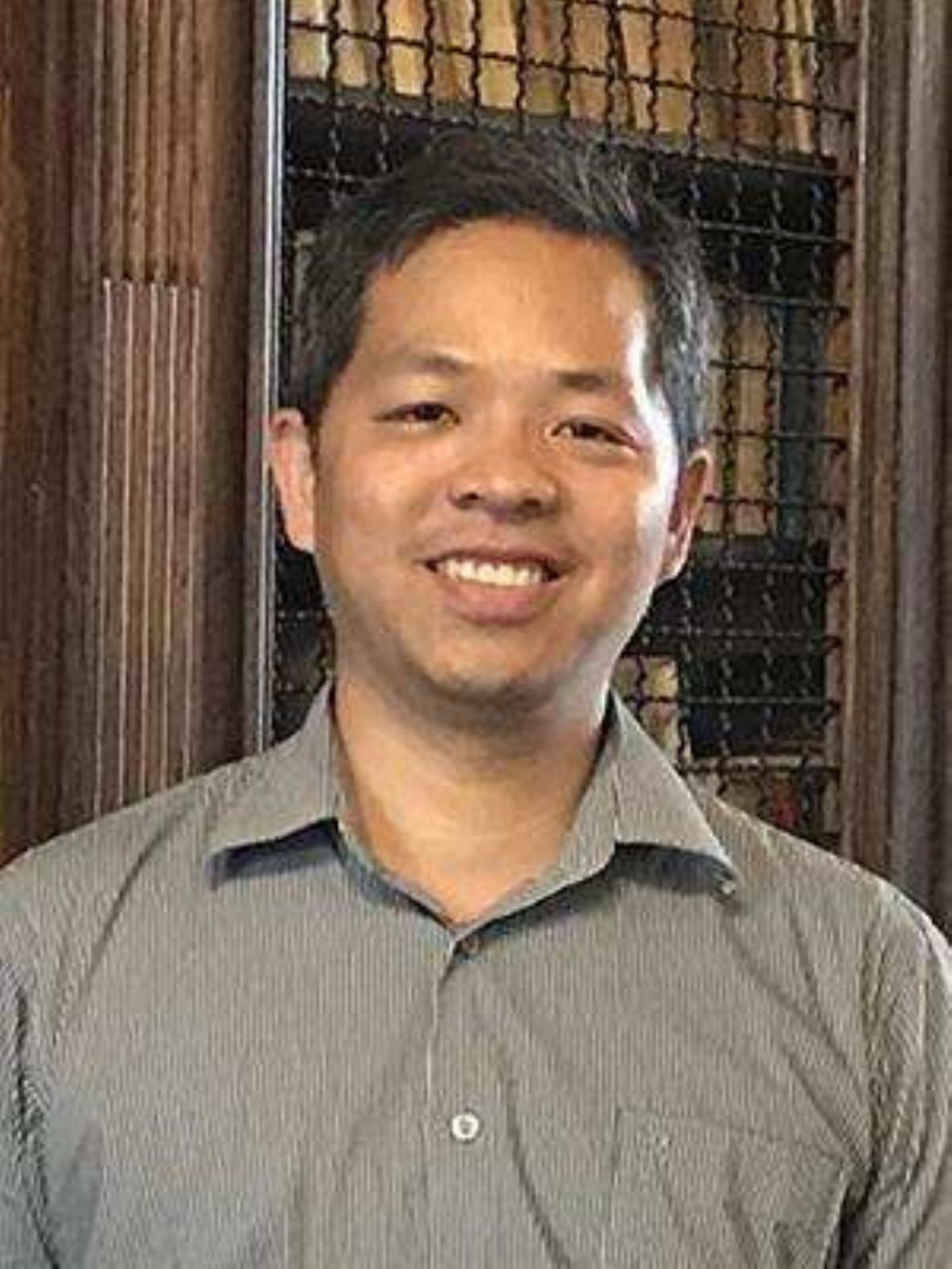}}]{Wen-Huang Cheng received the B.S. and M.S. degrees in computer science and information engineering from National Taiwan University, Taipei, Taiwan, in 2002 and 2004, respectively, where he received the Ph.D. (Hons.) degree from the Graduate Institute of Networking and Multimedia in 2008. He is a Professor with the Institute of Electronics, National Chiao Tung University (NCTU), Hsinchu, Taiwan, where he is the Founding Director with the Artificial Intelligence and Multimedia Laboratory (AIMMLab). Before joining NCTU, he led the Multimedia Computing Research Group at the Research Center for Information Technology Innovation (CITI), Academia Sinica, Taipei, Taiwan, from 2010 to 2018. His current research interests include multimedia, artificial intelligence, computer vision, machine learning, social media, and financial technology. He has received numerous research and service awards, including the 2018 MSRA Collaborative Research Award, the Outstanding Reviewer Award of 2018 IEEE ICME, the 2017 Ta-Yu Wu Memorial Award from Taiwan’s Ministry of Science and Technology (MOST), the 2017 Significant Research Achievements of Academia Sinica, the 2016 Y. Z. Hsu Scientific Paper Award, the Outstanding Youth Electrical Engineer Award from the Chinese Institute of Electrical Engineering in 2015, the Top 10\% Paper Award from the 2015 IEEE MMSP, the K. T. Li Young Researcher Award from the ACM Taipei/Taiwan Chapter in 2014. He is APSIPA Distinguished Lecturer.}
\end{IEEEbiography}
%\vspace*{-3\baselineskip}
%\vspace{-2.5cm}
\vskip -2\baselineskip plus -1fil

\begin{IEEEbiography}[{\includegraphics[width=1in,height=1.25in,clip,keepaspectratio]{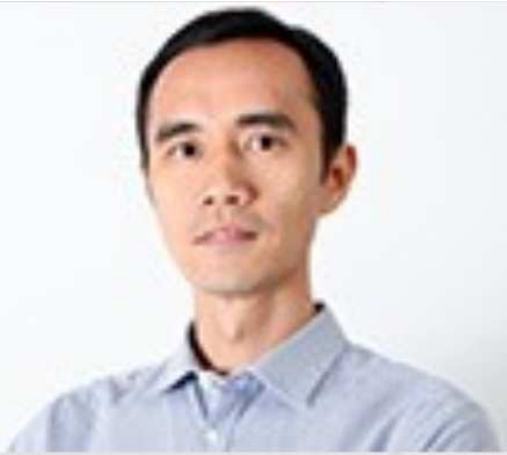}}]{ Neng-Hao Yu received Ph.D. degree in the Graduate
Institute of Networking and Multimedia from National Taiwan University, Taipei, Taiwan, in 2011. He is an Assistant Professor with the Department of Design, National Taiwan University of Science and Technology (NTUST), Taipei, Taiwan. Before joining NTUST, he created and directed the Innovative User Interface Lab (IUI Lab) at the Department of Computer Science, National Chengchi University (NCCU), Taipei, Taiwan, as an Assistant Professor from 2011 to 2018. His current research interests include human-computer interaction, user experience design, virtual reality, and multimedia technology. He has been a reviewer and chair for a wide range of international conferences in the field such as CHI, UIST, MobileHCI, Siggraph Asia and ChineseCHI.
}
\end{IEEEbiography}
\vskip -2\baselineskip plus -1fil

\begin{IEEEbiography}[{\includegraphics[width=1in,height=1.25in,clip,keepaspectratio]{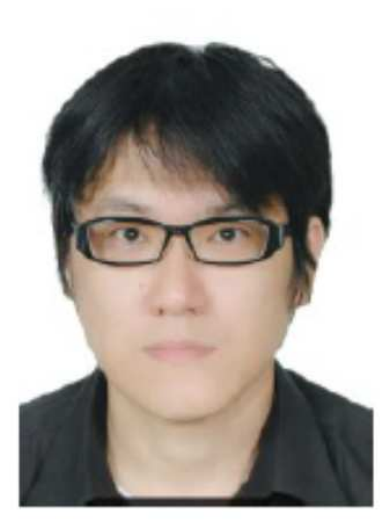}}]{Yu Tsao (Senior Member, IEEE) received the B.S. and M.S. degrees in electrical engineering from National Taiwan University, Taipei, Taiwan, in 1999 and 2001, respectively, and the Ph.D. degree in electrical and computer engineering from the Georgia Institute of Technology, Atlanta, GA, USA, in 2008. From 2009 to 2011, he was a Researcher with the National Institute of Information and Communications Technology, Tokyo, Japan, where he engaged in research and product development in automatic speech recognition for multilingual speech-to-speech translation. He is currently a Research Fellow (Professor) and Deputy Director with the Research Center for Information Technology Innovation, Academia Sinica, Taipei, Taiwan. His research interests include assistive oral communication technologies, audio coding, and bio-signal processing. He is currently an Associate Editor for the IEEE/ACM TRANSACTIONS ON AUDIO, SPEECH, AND LANGUAGE PROCESSING, IEEE SIGNAL PROCESSING LETTERS, and IEICE transactions on Information and Systems. He received the Academia Sinica Career Development Award in 2017, National Innovation Awards in 2018, 2019, and 2020, Future Tech Breakthrough Award 2019, and Outstanding Elite Award, Chung Hwa Rotary Educational Foundation 2019–2020.}
\end{IEEEbiography}
\end{document}